\documentclass{article}

\usepackage[left=2.54cm, right=2.54cm, top=2.54cm, bottom=2.54cm]{geometry}

\usepackage{longtable}
\usepackage{subcaption}
\usepackage{graphicx}
\usepackage{amsmath}
\usepackage{amssymb}
\usepackage[square, numbers]{natbib}
\usepackage[T1]{fontenc}
\usepackage{babel}

\usepackage{url}
\usepackage{hyperref}
\hypersetup{
    colorlinks,
    citecolor=blue,
    filecolor=blue,
    linkcolor=blue,
    urlcolor=blue
}

\DeclareMathOperator{\cond}{\vert}

\usepackage{booktabs}
\usepackage{setspace}

\title{Exploring the limits of Hierarchical World Models in Reinforcement Learning}

\author{
Robin Schiewer \\ \texttt{robin.schiewer@ini.rub.de} \\ \textit{Institute for Neural Computation, Department of Computer Science} \\ \textit{Ruhr University Bochum, Germany}
\and
Anand Subramoney \\ \texttt{Anand.Subramoney@rhul.ac.uk} \\ \textit{ Department of Computer Science} \\ \textit{Royal Holloway University of London, United Kingdom}
\and
Laurenz Wiskott \\ \texttt{laurenz.wiskott@ini.rub.de} \\ \textit{Institute for Neural Computation, Department of Computer Science} \\ \textit{Ruhr University Bochum, Germany}
}

\date{}

\begin{document}

\maketitle

\begin{abstract}
Hierarchical model-based reinforcement learning (HMBRL) aims to combine the benefits of better sample efficiency of model based reinforcement learning (MBRL) with the abstraction capability of hierarchical reinforcement learning (HRL) to solve complex tasks efficiently.
While HMBRL has great potential, it still lacks wide adoption.
In this work we describe a novel HMBRL framework and evaluate it thoroughly. 
To complement the multi-layered decision making idiom characteristic for HRL, we construct hierarchical world models that simulate environment dynamics at various levels of temporal abstraction. 
These models are used to train a stack of agents that communicate in a top-down manner by proposing goals to their subordinate agents. 
A significant focus of this study is the exploration of a static and environment agnostic temporal abstraction, which allows concurrent training of models and agents throughout the hierarchy. 
Unlike most goal-conditioned H(MB)RL approaches, it also leads to comparatively low dimensional abstract actions.
Although our HMBRL approach did not outperform traditional methods in terms of final episode returns, it successfully facilitated decision making across two levels of abstraction using compact, low dimensional abstract actions. 
A central challenge in enhancing our method's performance, as uncovered through comprehensive experimentation, is model exploitation on the abstract level of our world model stack. 
We provide an in depth examination of this issue, discussing its implications for the field and suggesting directions for future research to overcome this challenge. 
By sharing these findings, we aim to contribute to the broader discourse on refining HMBRL methodologies and to assist in the development of more effective autonomous learning systems for complex decision-making environments.
\end{abstract}

\section{Introduction}

In the past decade, deep reinforcement learning (RL) has demonstrated impressive success in complex sequential decision making problems \citep{mnih2013playing, lillicrap2019continuous, schulman2017proximal, haarnoja2018soft, silver2017mastering, vinyals2019grandmaster}.
However, deep RL is infamously known for its high demand for training data in order to perform well.
Model-based RL (MBRL) holds the potential to drastically reduce the required amount of training samples by integrating information from past environment interactions into a coherent model of the environment \citep{atkeson1997comparison, deisenroth2011pilco, chua2018deep, hafner2019learning}, often referred to as world model.
Going far beyond the capabilities of a simple replay memory, this model can then be used to produce synthetic training data for the agent.
Moreover, dynamics models can be used to guide exploration \citep{svidchenko2021maximum, kauvar2023curious} or implement risk-aware behaviour in agents \citep{yu2021riskaware}.

In an effort to scale RL to more complex and diverse problem domains, long term credit assignment and the ability to decompose tasks into subtasks became more relevant.
Hierarchical RL (HRL) has been proposed as a way to improve RL in both of these aspects. 
It divides the RL problem into distinct levels of increasing abstraction \citep{fikes1972learning, korf1985learning, schmidhuber1991neural, sutton1995modeling, sutton1999between, botvinick2014modelbased, vezhnevets2017feudal}.
The implementation of HRL usually results in a temporal or conceptual abstraction on the higher levels of the hierarchy, which alleviates the problem of long term credit assignment and can improve generalisation as well as exploration capabilities of the agent \citep{singh1992reinforcement, li2017efficient, florensa2017stochastic, xie2021latent}.
Furthermore, if a sequential decision making problems features repetitive structure, it can potentially be solved more efficiently with HRL.
By dividing behaviour into a lower level of abstraction that specialises on the various repetitive sub components of the problem and a higher level of abstraction that controls which lower level is active at a given time, problems consisting of arbitrary combinations of the sub components could be solved \citep{singh1992reinforcement, krishnamurthy2016hierarchical, florensa2017stochastic, xie2021latent}.
The central idea is essentially to have higher levels instruct lower levels of \textit{what} to do but not \textit{how} to do it. 
A popular theoretical framework for abstracting a sequential decision making problem is the options framework proposed by \citet{sutton1999between}.
It fixes various action sequences in place and designates them as new macro actions or ``options'' that have a temporally extended runtime. 
Over time, the definition of options has become increasingly diverse and multi-faceted.
Various interpretations of the options framework go beyond the notion of fixed action sequences and consider whole policies as options that can be orchestrated by a macro policy. 
In this context, the concept of options seamlessly transitions into the concept of skills, which are more versatile behavioural primitives like ``open a door'' or ``navigate to a specific object''.
Many approaches ranging from employing handcrafted options or skills to their automatic discovery have peen proposed over the years \citep{sutton1999between, mcgovern2001automatic, isahi2002qcut, simsek2004using, simsek2005identifying, konidaris2009skill, daniel2013autonomous, gregor2016variational, ramesh2019successor} and it can be argued that most, if not all, HRL methods make use of these two concepts in one way or the other.
However, reliably identifying a problem's subcomponents and crafting meaningful abstractions that allow flexible and powerful solution strategies is still a topic of ongoing research \citep{pateria2021hierarchical}.

At the intersection of HRL and MBRL lies hierarchical MBRL (HMBRL) \citep{singh1992reinforcement, li2017efficient, pertsch2020long, xie2021latent, hafner2022deep, mcinroe2023learning}.
It aims to combine the benefits of better sample efficiency with the strong suits of HRL, which are mitigating the long-term credit assignment problem, improving exploration and exploiting environment compositionality.
Unlike HRL, HMBRL involves the usage of a world model that may be hierarchical itself. 
Inspired by natural intelligence \citep{lee2003hierarchical, botvinick2014modelbased}, using a world model is considered a promising research direction to achieve more capable and intelligent autonomous agents \citep{lecun2023apath}.
However, the HMBRL landscape is less densely populated compared to model-free HRL, as combining the paradigms of HRL and MBRL produces highly complex and difficult to control systems.

One important detail in H(MB)RL is the decision of whether the length of a sub sequence, i.e. the temporal abstraction, should be variable or fixed.
While the former choice promises greater flexibility and is arguably more in line with intuition, it poses the additional problem of deciding when to start and end a particular option or low level world model.
If not properly regularised, approaches with variable-length temporal abstraction tend to degenerate to either one-step (i.e.\ no) temporal abstraction or to abstractions that span the whole sequence \citep{bacon2016optioncritic, alexander2016strategic, vezhnevets2017feudal, kim2019variational}, which is both clearly undesired. Finding expressive and diverse temporal abstractions without manual intervention or the injection of domain specific knowledge is still an open research question.

In this work, we present an HMBRL approach that consists of a hierarchical world model accompanied by a hierarchy of agents that are trained with simulated data from the world model.
We perform temporal abstraction of the training data in a static fashion independent of any agent behaviour, which enables concurrent training of all model and policy levels.
Furthermore, this sidesteps a moving target problem that can occur if higher levels of the model/policies are built on top of the capabilities of lower level models/policies, which tend to change over the course of training \citep{nachum2018data}.
We equip our higher world model levels with abstract actions that neither directly correspond to options nor are they goals for lower level policies. This results in far lower dimensional abstract actions compared to other goal-conditioned H(MB)RL approaches in the field.
As a proof of concept, we demonstrate that our approach works with a two level hierarchy and is able to generate hierarchical world models without using any domain knowledge or handcrafted abstractions.
We analyse our approach and provide insights on HMBRL, highlight crucial design decisions, and describe potential stumbling blocks that may arise in the building of HMBRL algorithms.

\section{Methods}

In this section, we describe the problem class we consider and detail our algorithm and its components. 
Specifically, we present the inner workings of our hierarchical world model, the agent that is trained via word model simulations and the process we use to temporally abstract the training data for the model. 
Additionally, we discuss the issue of model exploitation, where it can occur and possible strategies to combat it.  

\subsection{Problem Setup}
We consider episodic partially observable Markov decision processes (POMDPs) that are defined by a 7-tuple $\{S, A, R, T, \Omega, O, \gamma \}$.
$S$ represents the set of possible states of the MDP and $A$ denotes the set of actions available to navigate between them. Both can be finite or infinite sets, resulting in discrete or continuous state and action spaces. In this work, we focus on continuous spaces.
$R: S \times A \rightarrow \mathbb{R}$ is the reward function that assigns a scalar value to every state action pair.
It provides feedback about the quality of the most recent action in the most recent state.
$T$ is a set of conditional transition probabilities $T(s' | s, a)$ that define the probability to transition from any $s \in S$ to any other $s' \in S$ when performing $a \in A$.
$\Omega$ is a set of observations $o$ and $O$ is an accompanying set of conditional probabilities $O(o|s)$ that define the probability of $o \in \Omega$ occurring given $s \in S$.
The discount factor $\gamma \left[0, 1 \right)$ implements a trade-off between immediate and future rewards when the episode return $R = \sum_t \gamma^t r_t$, a measure of overall behaviour quality, is computed from the individual time step's rewards. 
Given a problem that can be formulated as a POMDP, the goal of reinforcement learning (RL) is to find a behaviour policy $\pi(s)$ that yields the highest possible return. 

\subsection{World Model}\label{sec:world_model}

For each level of our hierarchical world model, we use a variant of dynamical variational autoencoder (VAE) \citep{girin2020dynamical} known as the recurrent state space model (RSSM) \citep{hafner2019learning}. 
We largely follow the setup and training procedure of \citet{hafner2019learning} per model level. The RSSM is a versatile model capable of representing stochastic as well as deterministic environments. 
As the RSSM can adaptively accumulate information from past time steps if necessary, it can learn a Markovian transition model even for partially observable environments \citep{hafner2020dream}. 
Moreover, it has been demonstrated to perform well for environments with both large (e.g. image) and compact (e.g. proprioceptive) observation spaces \citep{xie2021latent}.

Being a generative model, the RSSM can produce sequences that mimic real environment interactions. To ground the model simulation, it has to be initialised with a few steps of data from the real environment, which is called a \textit{warm-up} or \textit{burn-in} period. 
After the warm-up period, the simulation is performed iteratively by repeatedly providing actions to the world model and letting it perform new simulation steps one by one, which is called a rollout. Thereby, the world model can operate in two modes: Closed loop and open loop. Closed loop rollouts continue the burn-in period for the duration of the whole rollout, which means the world model can base every simulation step on information from the real environment and essentially performs exclusively one-step predictions. In contrast to that, open loop rollouts rely on the internal state of the model to generate  simulation steps after the burn-in period and don't use any environment data after it.  

\begin{figure}
    \centering
    \includegraphics[width=\textwidth]{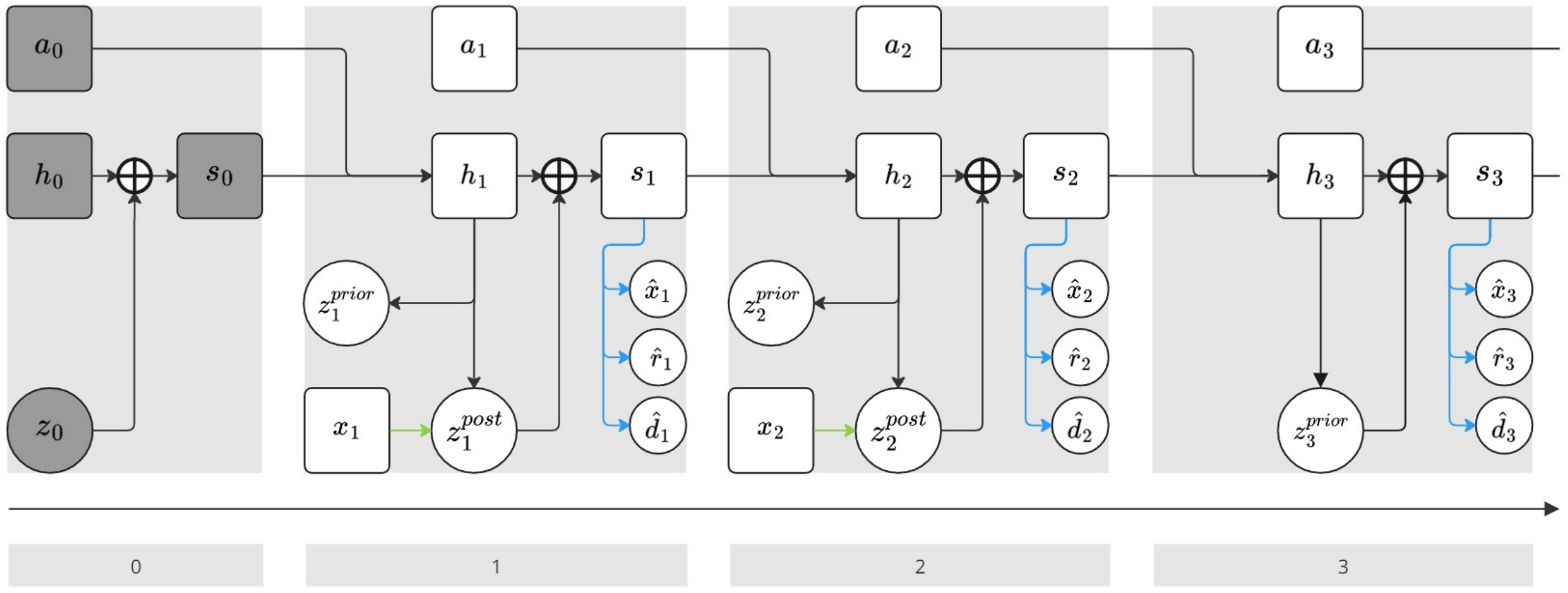}
    \caption{RSSM rollout graph for three time steps and one step of initial padding, time steps are grouped by grey boxes and indicated at the bottom for clarity. Squares indicate deterministic and circles indicate stochastic variables, $\oplus$ denotes concatenation and the shaded 0th time step represents initial zero padding. The RSSM state $s_t$ is shown as deterministic variable to emphasize that Equations~\ref{eqn:rssm_state_cl} and \ref{eqn:rssm_state_ol} are a deterministic operation once $z_t$ has been sampled. Ground truth information $x_t$ enters the RSSM via the encoder (green arrow), which implies $z_t^{post}$ is used for the next RSSM state. 
    $x_t$ refers to the observation $o_t$ for the level zero RSSM and to the goal $g_t$ otherwise. 
    If no ground truth information is available, $z_t^{prior}$ can fill the role of $z_t^{post}$ as shown in time step 3. 
    During model training, $x_t$, $r_t$ and $d_t$ are reconstructed to $\hat{x}_t$, $\hat{r}_t$ and $\hat{d}_t$ via their decoders (blue arrows). To provide a strong training signal for the world model, they are compared to the ground truth $x_t$, $r_t$ and $d_t$. The third step is open loop, which means $\hat{x}_t$, $\hat{r}_t$ and $\hat{d}_t$ are reconstructed from $z_t^{prior}$.}
    \label{fig:rssm}
\end{figure}

At its core, the RSSM contains a recurrent neural network (RNN) to deterministically capture information from past time steps and a VAE to represent uncertainty or stochasticity in the environment.
The concatenated RNN state $h$ and VAE encoding $z$ constitute the full RSSM state $s$, which can be used to decode rewards $r$, the probability of a trajectory terminating $d$, and observations $o$ on the lowest level or goals $g$ on all other levels.
Importantly, the prior $p$ and approximate posterior $q$ of the VAE are Gaussian distributions with mean vectors and diagonal covariance matrices parameterised by separate neural networks. They fill different roles depending on whether the RSSM is used for open or closed loop rollouts. The observation decoder may use a deconvolutional neural network in case of image observations to parameterise a diagonal Gaussian, whereas the goal decoder uses a regular multi-layer perceptron since the goals are always vectors. Both the reward and terminal models are additional decoder heads, with the reward decoder representing a univariate Gaussian and the terminal decoder a Bernoulli distribution. The RSSM can be described by the following set of equations:
\begin{align}
    \textrm{Deterministic state model:}\quad & h_t = f_\theta(h_{t-1}, s_{t-1}, a_{t-1}) \\
    \textrm{Representation model:}\quad & z_t^{post} \sim q_\theta(z_t^{post} \cond h_t, x_t) \\
    \textrm{Dynamics model:}\quad & z_t^{prior} \sim p_\theta(z_t^{prior} \cond h_t) \\
    \textrm{RSSM state (closed loop):}\quad & s_t = h_t \oplus  z_t^{post} \label{eqn:rssm_state_cl}\\
    \textrm{RSSM state (open loop):}\quad & s_t = h_t \oplus z_t^{prior} \label{eqn:rssm_state_ol}\\
    \textrm{Observation/goal decoder:}\quad & \hat{x}_t ~\sim q_\theta^x(\hat{x}_t \cond s_t) \label{eqn:rssm_obsgoal_decoder}\\
    \textrm{Reward decoder:}\quad & \hat{r}_t ~\sim q_\theta^r(\hat{r}_t \cond s_t) \\
    \textrm{Terminal decoder:}\quad & \hat{d}_t ~\sim q_\theta^d(\hat{d}_t \cond s_t),
\end{align}
where $\oplus$ represents the vector concatenation operator and the subscript $\theta$ indicates that all components are backed by trainable neural networks with the combined parameter vector $\theta$. The placeholder variable $x_t$ refers to observations $o_t$ for the lowest level RSSM and to goals $g_t$ for all other levels. Figure~\ref{fig:rssm} provides an illustration of a 3-step rollout with the RSSM and a burin-in period of 2 steps.

A single prediction step of the RSSM works as follows:
The last RSSM state $s_{t-1}$ and action $a_{t-1}$ are fed to the RNN to produce a new deterministic state component $h_t$.
In closed loop mode, $h_t$ is combined with the current $x_t$ and fed to the approximate posterior network $q$ to form the current stochastic state component $z_t^{post}$. Having access to $h_t$ equips the approximate posterior network with information about the past and, at the same time, it serves as an inlet for new information in form of the current observation $o_t$ or goal $g_t$ shown as $x_t$ in Figure~\ref{fig:rssm}. This is why $q$ is named the \textit{representation model} component.
In open loop mode, the VAE's prior $p$ is used instead of $q$ to generate the stochastic state component $z_t^{prior}$. While having information about past time steps, $p$ does not obtain the current observation and can thus not ground its predictions. Instead, is has to predict the dynamics of the environment as accurately as possible solely from the last state and action, which is why it is called the \textit{dynamics model} component.
During training, predictions of $p$ as well as $q$ are generated and minimisation of the KL divergence between them ensures that the dynamics model is capable of performing useful and accurate rollouts. Thus, in contrast to a regular VAE the prior is not a static uniform Gaussian distribution, but is learned as well.
Using the RSSM state $s_t$, the \textit{observation/goal decoder}, \textit{reward decoder}, and \textit{terminal decoder} components predict the current time step's observation $o_t$ or $g_t$, reward $r_t$, and the probability of the trajectory terminating $d_t$, respectively.
We use the hat notation to indicate sampled estimates for observations, rewards and terminal flags to differentiate them from their ground truth counterparts. The RSSM is trained via minimising the evidence lower bound objective (ELBO), which yields the following loss function:

\begin{align}\label{eqn:rssm_reconstruction_lossses}
    \mathcal{L}_t^{o} & = - \ln q_\theta(\hat{o_t} \cond s_t) & \mathcal{L}_t^{r} & = - \ln q_\theta(\hat{r}_t \cond s_t) & \mathcal{L}_t^{d} & = - \ln q_\theta(\hat{d}_t \cond s_t) 
\end{align}
\begin{align}\label{eqn:rssm_kl_loss}
    \mathcal{L}_t^{KL} & = - D_{KL}\left[ q_\theta(z_t \cond h_t, o_t) \,\Vert\, p_\theta(z_t \cond h_t) \right]
\end{align}
\begin{align}
    \mathcal{L} &= \mathbb{E}_{q_\theta(z_{1:T} \cond a_{1:T}, o_{1:T})} \left[ \frac{1}{T} \sum_{t=1}^T \beta \mathcal{L}_t^{KL} + \mathcal{L}_t^{o} +  \mathcal{L}_t^{r} +  \mathcal{L}_t^{term} \right],
\end{align}
whereas $x_{1:T} = x_1, x_2, ..., x_T$ and $\beta$ is a coefficient that trades off the KL divergence loss term (Equation~\ref{eqn:rssm_kl_loss}) with the reconstruction loss terms (Equations~\ref{eqn:rssm_reconstruction_lossses}). 
A schematic overview of the RSSM is shown in Figure~\ref{fig:rssm}.
Note that Equation~\ref{eqn:rssm_kl_loss} actually represents the KL divergence between the representation and the dynamics model for only one time step, implying that the RSSM is trained as a one step predictive model.
However, the RSSM is capable of performing multi-step open loop rollouts (i.e. stacking repetitive applications of the dynamics model) of a quality high enough to use them for agent training. 
As with other world model architectures, it accumulates error that grows with the rollout length though \citet{nagabandi2017neural, hafner2019learning, hafner2020dream, janner2021trust}, which limits the reasonable maximum length of rollouts to around 15 to 20 time steps depending on the complexity of the simulated environment.
When the RSSM was first introduced, \citet{hafner2019learning} proposed an additional loss that took into account the KL divergence between priors from multi step rollouts and one step posteriors to increase robustness of the open loop rollouts and called this technique \textit{latent overshooting}. 
They found it to be beneficial when using the RSSM in conjunction with random shooting algorithms like CEM for model predictive control. 
However, later they noted it became obsolete after switching to an actor-critic agent that is trained on the RSSM simulations \citep{hafner2020dream}. 
We did not find any performance improvement using latent overshooting either and thus omit it for simplicity.

Using the RSSM, our goal is to craft a hierarchy of stacked world models $M^l$, $l \in \{0, 1, ..., L\}$, where higher levels represent the environment at an increasingly coarse temporal resolution, i.e.\ the higher the level the larger the temporal stride of the model. This stack of world models can be interpreted as a collection of interconnected POMDPs which can in principle be solved individually by arbitrary RL algorithms. The training data for the individual world model levels is produced from ground truth data and world model states as described in Section~\ref{sec:temporal_abstraction} and we establish the connection between adjacent world models via intermediate goals that associate higher and lower level model states with each other. Entering a specific state $s^l$ in $M^l$ spawns a goal that refers to a world model state $s^{l-1}$ in $M^{l-1}$, which effectively links the two levels with each other. Given two states $s_a^l$ and $s_b^l$ in $M^l$, their associated goals $s_a^{l-1}$ and $s_b^{l-1}$ can be interpreted as abstract descriptions of places in $M^{l-1}$, whereas $M^l$ is indifferent about the specifics of how to get from $s_a^{l-1}$ to $s_b^{l-1}$.

\begin{figure}
    \begin{minipage}{0.815\linewidth}
        \centering
        \includegraphics[width=\linewidth]{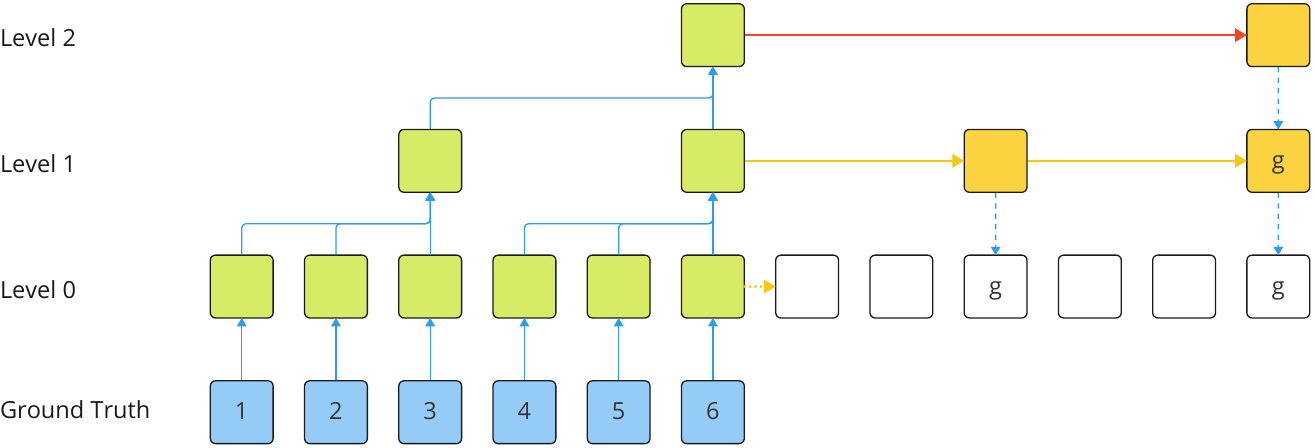}
    \end{minipage}\hfill
    \begin{minipage}{0.155\linewidth}
        \centering
        \includegraphics[width=\linewidth]{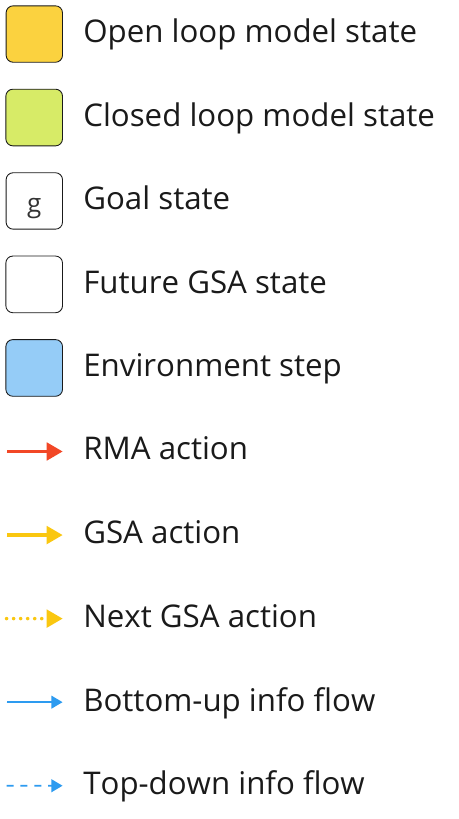}
    \end{minipage}
    \caption{An overview of our method using a 3 level model with temporal abstractions of 3 steps on level 1 and 2 steps on level 2. The level 0 model does not perform any temporal abstraction and simulates the ground truth data from the environment at full temporal resolution. Each node represents a time step on the respective level. In this example, as shown by the blue nodes, there are 6 steps of ground truth data available. Green nodes on the model levels indicate closed loop time steps that imply the model has been grounded via data from the real environment for these steps. The bottom-up information flow for model grounding is visualised by the blue arrows. Yellow nodes indicate open-loop steps that are not grounded. On Level 2, the RMA action (red arrow) generated a new open loop model state that spawned a goal, marked with the letter ``g'' on level 1. By navigating to this goal, the level 1 GSA executed two actions symbolised by the yellow arrows and produced two goals for the level 0 GSA. The GSA on level 0 has now two goals and a budget of 3 actions per goal to navigate towards them. Its immediate next action, which will also be executed in the real environment to collect data for time step 7, is indicated by the dotted yellow arrow. As soon as the level 0 GSA depleted its total action budget, the 6 data from the 6 newly collected environment steps will be used to ground the model levels 0 to 2 and equip the RMA on level 2 with up-to-date information for choosing the next action.} 
    \label{fig:model_diagram}
\end{figure}

On each model level, we employ two different agents, one for maximising the environment reward on this level and on for following goals given from the level above. We refer to them as the \textit{reward maximising agent} (RMA) and \textit{goal seeking agent} (GSA), respectively. While the need for GSAs on all but the top most level is obvious, it can be argued that only a single RMA at the top most level is required. However, in its current form our approach relies on the RMAs on the lower levels for decision making during the initial time steps. At the beginning of the environment interaction, depending on their level and temporal stride not all world model levels can be initialised with ground truth data right away. This is simply due to not enough actual environment steps being available to perform the temporal abstraction process that is necessary to preprocess the ground truth data before higher level models can ingest it. In that phase, we use the highest possible RMA that already has an initialised world model and resort to GSAs on all levels below. An overview of our architecture is shown in Figure~\ref{fig:model_diagram}.

\subsection{Temporal Abstraction}\label{sec:temporal_abstraction}

Given a ground truth trajectory $\tau$, each world model $M^l$ in the hierarchy is trained on an increasingly coarser version of $\tau$, referred to as $\tau^l$. 
Thereby, $M^0$ is a special case as it models the environment at its original temporal resolution. 
Hence, it receives the unmodified trajectory $\tau^0 = \tau$. 
For all $M^l$ with $l > 0$, $\tau^l$ is supplemented by world model states gained by closed loop rollouts using $\tau^{l-1}$ with $M^{l-1}$. Those world model states are required to semantically connect $M^l$ with $M^{l-1}$ (c.f. Section~\ref{sec:world_model}). We achieve temporal abstraction by dividing $\tau^l$ into subtrajectories of length $k^l$ and applying filters to each chunk\footnote{If $\tau$ is not evenly divisible by $k$, it is padded and the operations presented below use masks to effectively consider only the non-padded portion of the sequence.}.
\\
\textbf{Observations:} 
Since we construct latent space world models, the levels $l > 0$ do not predict any observations, rendering the ground truth observations irrelevant for all models except $M^0$. Hence, no temporal abstraction is necessary for observations.
\\
\textbf{Goals:} 
For $l = 0$, no goals need to be produced as no lower level model exists. Conceptually, for higher levels the goals can be thought of as filling the role of the observations that are only predicted by $M^0$. Here, the last time step's model state per chunk is selected:
\begin{equation}
    g_i^l = s_{ik^l}^{l-1}
\end{equation}
\\
\textbf{Rewards:} For all levels, the average reward per chunk $i$ is calculated:
\begin{equation}\label{eqn:temporal_abstraction_r}
    r_i^l = \frac{1}{k^l} \sum_{j=(i-1)k^l}^{ik^l} r_j^{l-1}
\end{equation}
\\
\textbf{Terminals:} For all levels, the maximum terminal flag per chunk $i$ is calculated:
\begin{equation}
    d_i^l = \max_j \{d_j^{l-1}\}, \quad j \in \{(i-1)k^l + 1, \ldots, ik^l\}
\end{equation}
\\
\textbf{Actions:} We evaluated various ways to obtain low dimensional, compressed abstract actions from primitive action sequences. The central goal here was to achieve abstract actions that are lower in dimension than the sequence of concatenated primitive actions.
\begin{enumerate}
    \item Compress the action sequence chunks $\{a_j^{l-1}\}, \; j \in [(i-1)k^l + 1, ik^l]$ into a lower-dimensional representation $a_i^l$ using a VAE. This allows us to obtain detailed and versatile high level actions that feature a similar level of expressivity compared to the low level sequences.
    \item Cluster action sequence chunks via k-means, whereas centroids are updated via moving averages. This way, the higher level actions become discrete, which greatly helps to avoid the potential exploration challenges that may arise from high dimensional continuous abstract action spaces.
    \item Use a fixed random projection to map low level action sequences to high level actions. This retains the expressivity of a continuous action space and makes the temporal abstraction process completely static over the whole course of the training.
\end{enumerate}

We found discrete abstract actions through action sequence clustering to generally perform poorly. Our hypothesis is that the strong compression of the action sequences from level 0 to level 1 introduces a high level of stochasticity on level 1, which impairs the learning process of the RMA and reduces expressivity of the world model. Thus, we chose the autoencoder method for generating abstract actions as described in Section~\ref{sec:temporal_abstraction} and adhere to the equations proposed in the same section otherwise. i.e.\ for the observations, rewards, terminal flags and goals. In an attempt to mitigate model exploitation as explained in Section~\ref{sec:model_exploitation}, we use the beta VAE variation for generating abstract actions from lower level action sequences. We chose it due to its simplicity and since it is only a small modification from the regular action autoencoder setting. 

\begin{figure}[h]
    \begin{center}
    \includegraphics[width=1.0\textwidth]{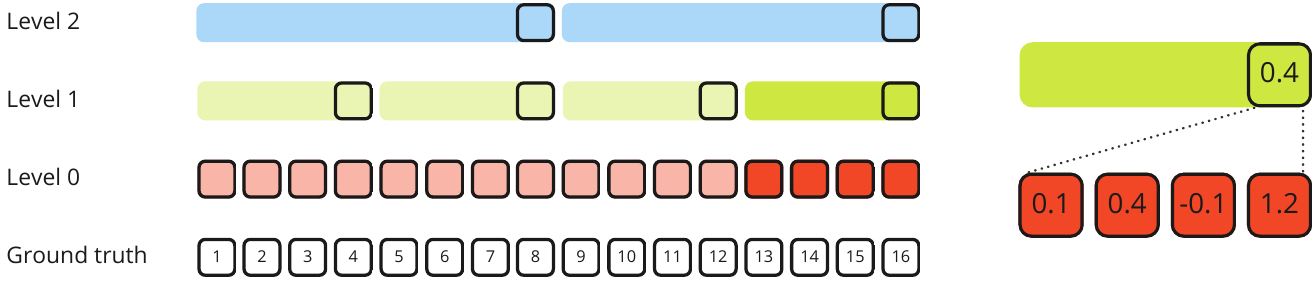}
    \caption{Trajectories from a hierarchical world model with three levels, each time step's data (observation, reward, terminal flag, model state, etc.) is represented by a square. 
    The shaded areas indicate how many real world time steps the individual model levels span. 
    Level 0 (red) is a special case that models the true MDPs dynamics with a temporal stride of $k^0=1$. 
    Levels 1 (green) and 2 (blue) have a temporal stride of $k^1=4$ and $k^2=2$, respectively. 
    The reward abstraction process for the 4th step on level 1 (more saturated colouring) is exemplified in detail on the right. 
    The reward on the upper level is the mean of the lower level chunk ranging from steps 13 to 16. inclusive.}
    \label{fig:temporal_abstraction}
    \end{center}
\end{figure}

Irrespective of the choice of abstraction method for the actions, all approaches we tested were intended to avoid a moving target problem originating from the abstraction process having a dependence on the agent performance.
Consider the situation where the rewards on some level $l > 0$ depend on the GSA performance on level $l - 1$, specifically the rewards on level $l$ are calculated according to Equation~\ref{eqn:temporal_abstraction_r} from the individual rewards the GSA on level $l - 1$ obtains while trying to navigate to a given goal. This can potentially result in heavily varying action values on level $l$ as described by \citet{pateria2021hierarchical} due to the change in performance of the GSA on level $l - 1$ over the course of training. 
Specifically, the problem arises when the level $l - 1$ GSA is not yet trained well and thus can't navigate properly to a given goal, and the same goal may yield a notably different reward later on once the level $l - 1$ GSA did improve its goal navigation abilities. 
This makes it difficult for the level $l$ RMA to trust initial action values and represents a challenging exploration problem. The described abstraction scheme and a concrete example for the reward are illustrated in Figure~\ref{fig:temporal_abstraction}.

\subsection{Goal and Similarity Measure}

This section explores three critical decisions in designing goals and their similarity measures in our framework. 
In our approach, the goals enable agents on lower levels of the hierarchy to execute the orders of their superiors. They are derived from RSSM states and are essentially high-dimensional vectors with both deterministic and stochastic components. 
Measuring similarity is challenging due to the complex nature of these goals and the variations introduced by their stochastic elements. 
We discuss different similarity measures, the aspects of comparison, and the choice between relative and absolute goals to address these challenges effectively.

\textbf{What to compare?} We train our hierarchical world models to output lower level model states as goals, i.e. $M^l$ outputs goals in the state space of $M^{l-1}$. One option is to compare the target states i.e.\ goals produced by $M^l$ directly to states from $M^{l-1}$, as done in \citet{hafner2022deep, vezhnevets2017feudal}. This option worked reliably across our experiments, although for simpler environments we found comparing the reconstructed observations (described below) as done in \citet{nachum2018data} to result in slightly faster learning.
This might be due to the fact that the RSSM states can be high dimensional and hard to differentiate using a similarity measure such as the Euclidean distance or an inner product. We conjecture this is due to a more distributed representation being learned by the RSSM, where states corresponding to similar regions in latent space do not necessarily lie close to each other in latent space since the RSSM has more dimensions available to store information than are necessary. Yet, we can't simply reduce the state size, as there arises a dilemma when searching for the optimal dimension: Even when faced with a comparatively simple environment, the predictive power of the RSSM is strongly influenced by its capacity, i.e.\ the number of units in the RNN plus number of units in the VAE bottleneck. While increasing the capacity yields a lower prediction error and faster learning, the RSSM states may also become harder to compare with each other as outlined above.

Reconstructing level 0 environment observations from goals on level $l$ (or states on level $l-1$) works as follows: We employ the goal/observation decoder of $M^{l-1}$ (Equation \ref{eqn:rssm_obsgoal_decoder}), which is specifically trained to understand the possibly complex and distributed states to decode a lower level RSSM state or observation from it. 
It can be used on goals produced by $M^{l}$ (target states for $M^{l-1}$) to decode them into level $l-2$ states. 
This can be cascaded until we arrive at $M^0$, where we can reconstruct actual observations from the states. 
These observations can be compared for computing the similarity score.
An alternative option to comparing raw states or reconstructed observations is to compress the goals of $M^l$ and states of $M^{l-1}$ via a VAE in an effort to reduce redundancy in its representation. 
Although this option holds the potential to generalise better across a range of different environments, it adds more hyperparameters and yet another trainable component to the architecture that changes over time and might slow down training.
Beyond the options discussed here, \citet{nachum2019near} present an approach to learn near-optimal representations specifically for the purpose of goal-oriented navigation in HRL. However they work in a model-free HRL setting and can choose their goal representations as needed. In our case, the RSSM states need to work well in the context of their world model and other objectives are inevitably secondary to this.

\textbf{What similarity measure to use?} A straightforward approach is to compute the negative Euclidean distance between goals to assign a similarity score to them as \citet{nachum2018data} do. For a RSSM state $s$ and a goal $g$, this means: 
\begin{equation}\label{eqn:goal_similarity_mse}
    \mathcal{M}_{MSE}(s, g) = - \Vert g - s \Vert
\end{equation}
In contrast to that, \citet{hafner2022deep} propose the scaled dot product similarity measure:
\begin{equation}\label{eqn:goal_similarity_ndp}
    \mathcal{M}_{SDP}(s, g) = \frac{1}{m}g^T \frac{1}{m} s \quad \text{where} \quad m = \max(\Vert g \Vert, \Vert s \Vert).
\end{equation}
It is similar to the measure proposed in \citet{vezhnevets2017feudal} and we found a large variety of additional measures in the source code of \citet{hafner2022deep}, which suggests yet unexplored or untapped potential in the choice of the similarity measure. 
Following a different idea, \citet{liu2022searching} argue that the above methods expect a densely populated goal space, which may not be true in practice. Based on the assumption that the goal space is a Riemannian manifold curled up in latent space, goals with a small Euclidean distance can be far away from each other in the actual goal space. They propose a specific learning process for the goals, accompanied by a complex distance measure. However, their embedding approach introduces constraints that are not directly compatible with our world model architecture and opens up a new avenue of research that is beyond the scope of this work.

\textbf{Relative or absolute goals?} Due to simplicity in the abstraction and world model training procedures, our world model proposes absolute goals. Yet, we transform them before they are presented to the lower level agent. Given a goal $g_t^l$ proposed by $M^l$ and the current state $s_t^{l-1}$ of $M^{l-1}$, the transformed goal is $\overline{g}_t^{l} =  s_t^{l-1} - g_t^l$. This transformation produces relative goals, thereby following the findings of \citet{nachum2018data, zhang2020generating} who state that relative goals help the agent to generalise better. In contrast, \citet{vezhnevets2017feudal, hafner2022deep} use absolute goals, which we did not investigate in detail in this work.

\subsection{Agent Training}\label{sec:agent_training}

We use actor-critic agents as the RL algorithm of our choice and, as with our world models, largely follow the setup of \citet{hafner2019learning}. 
We train all agents using backpropagation, through the differentiable learned world models. 
The observation space of the agents on level $l$ is the state space of $M^l$, whereas the goal seeking agent (GSA) has an observation space twice as large as it additionally obtains a goal in form of a state $s^l$. 
Each agent features separate actor and critic networks with parameters $\psi$ and $\xi$, respectively. 
For clarity of notation, we will omit the hierarchy index $l$ from the actor and critic equations in the following, since all quantities refer to those in the same level of hierarchy. 
To obtain actions and state values for the reward maximising agent (RMA) on any level given a world model state $s_t$, we have:
\begin{align}
    a_t &\sim  p_\psi(a_t \cond s_t) \\
    V_\xi(s_t) & \approx \mathbb{E}_{p_\psi, p_\theta} \left[ \sum_{k \geq t} \gamma^{k-t} \prod_{k \geq j \geq t}(1 - \hat{d}_{j-1}) \hat{r}_k \right]
\end{align}
where $\gamma$ is a fixed discount parameter, $\hat{d}_t$ is the model estimate for the terminal flag that is used for discounting\footnote{Note: As $d_t = 1$ indicates an episode ending, we use $1 - \hat{d}_t$ for discounting. If a simulation ends in time step $t$ due to $\hat{d}_t$ being predicted close to 1, we don't want to mask the final reward in that step, but everything that may come after it in case the simulation continues. Thus we need to shift the time index of the terminal flags by one.} alongside $\gamma$ and $\hat{r}_k$ is the k-th reward after starting the simulation.

In a standard fashion, we train the RMA on each level $l$ to achieve maximum episode return in the simulations generated by $M^l$. We compute TD-$\lambda$ state values \citep{sutton2018reinforcement} to balance the bias of temporal difference against the variance of Monte Carlo REINFORCE \citep{williams2004simple} training targets. The recursive definition of the training targets is:
\begin{equation}\label{eqn:td_lambda_returns}
    V_t^\lambda = \hat{r}_t + \gamma \hat{d}_t
    \begin{cases}
        (1 - \lambda) v_\xi(s_{t+1}) + \lambda V_{t+1}^\lambda & \text{if} \quad t < H, \\
        v_\xi(s_{H}) & \text{if} \quad t = H.
    \end{cases}
\end{equation}            
Here, $H$ refers to the maximum length of the open loop rollouts used to train the agents. The critic is trained to minimise the MSE between the estimated state values and their TD-$\lambda$ targets:
\begin{equation}\label{eqn:critic_loss}
    \mathcal{L}_{\text{critic}} = - \mathbb{E}_{p_\psi, p_\theta} \left[ \sum_{t=1}^{H}\frac{1}{2} (v_\xi(s_t) - V_\lambda(s_t))^2 \right],
\end{equation}
whereas we stop gradient flow for the second term $V_\lambda(s_t)$ as usual. Since we can propagate gradients through the dynamics model and critic, the actor is directly trained via gradient descent on the TD-$\lambda$ value estimates of the states visited in the simulation. As an additional stability measure proposed by \citet{hafner2020dream}, we store for each agent a separate exponentially moving average (EMA) copy of its value network. During actor training, we choose for each state value the minimum of the live and the EMA critic. Thus, the main actor loss is:
\begin{equation}\label{eqn:actor_dyn_loss}
    \mathcal{L}_{\text{dynamics}} = - \mathbb{E}_{p_\psi, p_\theta} \left[ \sum_{t=1}^{H} V_\lambda(s_t) \right],
\end{equation}
where $v_\xi(s_t)$ from Equation~\ref{eqn:td_lambda_returns} is replaced by $\min(v_\xi(s_t), v_{EMA}(s_t)$. Note that although the parameters of the actor do not appear in Equation~\ref{eqn:actor_dyn_loss} directly, the TD-$\lambda$ returns are calculated from open loop rollouts obtained by letting the actor perform inside the world model. This makes them directly dependent on the policy network. As additional actor loss terms, we use a combination of epsilon-greedy exploration and reward shaping to solve the exploration-exploitation trade-off. Similar to \citet{haarnoja2018soft}, we encourage a high entropy of the action distribution to facilitate random behaviour and thus exploration in the absence of reward signals:
\begin{equation}\label{eqn:actor_entropy_loss}
    \mathcal{L}_{\text{ent}} = - \mathbb{E}_{p_\psi, p_\theta} \left[ \sum_{t=1}^{H} \mathrm{H}(a_t \cond s_t) \right].
\end{equation}
Since we have a world model available, we augment the reward with the kl-divergence between the rollout data produced by an up-to-date and an exponentially moving average (EMA) version of the world model:
\begin{equation}\label{eqn:actor_novelty_loss}
    \mathcal{L}_{\text{novelty}} =  - \mathbb{E}_{p_\psi, p_\theta} \left[ \sum_{t=1}^{H} - D_{KL}\left[ p_\theta(z_t \cond h_t)] \,\Vert\, p_{EMA}(z_t \cond h_t) \right] \right].
\end{equation}
This can be interpreted as a novelty reward that is higher for regions where the world model did recently change and represents an alternative to using ensembles of models for the same purpose \citep{chua2018deep}. The final actor loss is the combination of Equations~\ref{eqn:actor_dyn_loss}, \ref{eqn:actor_entropy_loss} and \ref{eqn:actor_novelty_loss}:
\begin{equation}
    \mathcal{L}_{\text{agent}} = \mathcal{L}_{\text{dynamics}} + \eta \mathcal{L}_{\text{ent}} + \mu \mathcal{L}_{\text{novelty}}
\end{equation}
with the individual loss term weights $\eta$ and $\mu$.

\subsection{GSA Training}

The GSA (goal seeking agent) training largely follows the RMA procedure, yet there are three key modifications. Firstly, observations are concatenated with the relative goal, i.e. $s'_t = s_t \oplus \overline{g}$. Secondly, rewards are not the actual world model's step rewards but calculated based on the similarity between the current world model state and the goal state via $\mathcal{M}_{MSE}(s, g)$ or $\mathcal{M}_{NCS}(s, g)$. Thirdly, the terminal flag the GSA receives is not the one of the environment, but similar to the reward calculated based on the similarity of the world model state and the goal.

Depending on whether a similarity measure operates in the positive regime as $\mathcal{M}_{NCS}(s, g)$ or the negative regime as $\mathcal{M}_{MSE}(s, g)$, a pathologic case can occur: If the GSA has access to the terminal flags simulated by its world model, i.e. in essence to the terminal states of the real environment, it may choose to ignore goals if such a terminal state is easier to reach than the current goal to avoid further penalties. This creates absorbing regions around terminal environment states in which the GSA will always navigate to those terminal states, which might be undesirable depending on the environment. If the goal reward is chosen to be positive and increasing the closer the agent gets to the goal, as it is the case in various related works, we found real world terminal states to become repulsive instead of attractive. This makes sense as the GSA ceases to obtain positive reward if the trajectory ends, so it avoids to enter terminal states. This avoidance translates to a large degree to situations where terminal states are the goal, resulting in suboptimal overall performance.
To avoid this, the third modification we propose is to completely prevent the GSA from accessing the terminal flag of the world model. This raises the question how to handle if a GSA happens to step into a terminal state by accident or while trying to reach a given goal. The simulation would have to carry on although the simulating world model was by definition never trained on transitions that transcend the episode ending. While we may expect a certain amount of generalisation from the world model, it is queried out of distribution in such cases. A reasonable strategy to avoid this problem could be to generate custom terminal flags for the GSA based on the reward signal. Ideally, the GSA trajectory should end once it reaches its goal. If we furthermore decide to always pick the start and goal for the GSA simulations from trajectories of our replay memory, we can make sure that the goals are always valid and reachable. Another argument in favour of letting GSA training rollouts end once the agent reaches its goal is the emergence of a jittering movement pattern around the goal that we found in navigation tasks when we didn't use terminal flags for the GSA. This pattern makes sense as the only option the agent has once it reaches its goal is to move around in the goal region to keep penalties low or proximity reward high, depending on the chosen reward scheme. However, we noted the tendency of this jittering, which is clearly an artefact of the GSA training method, to manifest in the collected data in the form of longer, lower reward trajectories. Since the quality of the collected data directly influences the training of the whole hierarchical model and all of its agents, it is relevant to preventing the jittering. We propose the following transformation of the rewards $\mathcal{M}_{MSE}(s, g)$ to generate terminal flags for the GSA:
\begin{equation}
    d_t = \sigma \left( \left( r_t + \frac{p + c}{2} \right) \frac{\sigma_{\max} - \sigma_{\min}}{p - c} \right),
\end{equation}
with $\sigma$ referring to the sigmoid function, $p$ to the reward value at which the probability of the episode end should start to become larger than zero, $c$ to the reward value at which the probability of the episode end should be one, $\sigma_{\max}$ to a maximum cutoff value and $\sigma_{\min}$ to a minimum cutoff value. The cutoff values reflect the idea that $\sigma$ will be effectively zero for values smaller than $\sigma_{\min}$ and effectively one for values larger than  $\sigma_{\max}$. We found $p = -0.01$, $c = -0.0001$, $\sigma_{\min} = -5.0$ and $\sigma_{\max} = 5.0$ to work well in our experiments but especially $p$ and $c$ might benefit from specific tuning for environments other than those we tested

\subsection{Model Exploitation}\label{sec:model_exploitation}

When learning a world model along with a behaviour policy, as is common practice in HMBRL, the model may be incorrect outside of the regions covered by training data from environment interaction \citep{chua2018deep}. 
Thus, the model depends on the data collection policy and \textit{vice versa} if the world model is involved in the agent training.
A common complication in this setting is a situation where the agent finds inaccuracies of the world model by accident and learns to exploit them for supposedly high return \citep{janner2021trust}. 
In reality, behaviour learned in this way will not carry over well to the real world where the model inaccuracy does not exist. 
Usually, in such a case the world model will be corrected once the behaviour learned in the model simulations is applied to collect new ground truth data from the real environment. 
These new set of trajectories, which will contradict the erroneous model simulations, will effectively repair precisely the model inaccuracies that were previously exploited by the agent. 
Over the course of training, competition between the agent and the model becomes less severe. 
On the one hand, the model exhibits less and less defects and on the other hand, the agent's behaviour becomes increasingly guided and situated in regions of the world model that are well covered by ground truth data. 
Note that even in a well working MBRL approach, the model may still be incorrect in regions that are seldom or never visited by the agent\footnote{Note that the situation might be different for random shooting algorithms like CEM, as they always start with a random sequence of actions that gets successively refined. This is likely to put more emphasis on global model correctness, because the CEM can provide any sequence of actions to the model irrespective of the simulation's starting point. Compared to that, a learned policy will quickly resort to much more guided and state-specific action sequences, i.e.\ it will ``learn to ask the questions that matter'' in terms of reward in the respective areas of the environment.}.

In our hierarchical world model scenario, we argue that model exploitation can happen depending on the abstraction process used for the actions. 
Specifically, the abstraction process from level $i \geq 0$ to $j = i + 1$ may produce action mappings $a^j$ that lie in a restricted region of the space of all possible actions on level $j$. 
However, the agent on level $j$ can propose actions outside of that region or where the abstracted actions have thin or no coverage. 
Even if the abstract action space is constrained, e.g. by limiting each dimension to have values between -1 and +1, this problem can still occur due to the continuous nature of the action space. 
In essence, some of the actions proposed by the agent on level $j$ may not be in the training data set of its world model, and so, the world model on level $j$ will produce wrong rollouts if it receives them as input. 
This provides the agent on level $j$ with the opportunity to exploit its world model with uncommon actions in order to obtain seemingly high rewards.
Even if over time the abstraction process spreads its mappings more evenly over the abstract action space, the abstract agent may adapt to this and learn new exploitative actions.
This effect arguably worsens if multiple unusual or out-of-distribution actions are fed to the model in a row, resulting in decreasingly common model states in each step.
We can't easily patch this flaw in the world model as long as we do not have direct control over the mapping resulting from the abstraction process. 
We explored the following ways to mitigate this problem:

\begin{itemize}
    \item \textbf{Discrete actions:} If we employ an abstraction process that produces discrete abstract actions, e.g.\ clustering or some form of binning, it can be ensured that for every element in the abstract action space on level $j$ there exists at least one counterpart in the space of action sequences on the level $i$. Thus, it becomes impossible for the agent on level $j$ to find actions that are not covered by ground truth data, which means over time all exploitable model inaccuracies can be repaired. A clear disadvantage of discrete actions is the limited expressivity if the action space on level $i$ is continuous or even if it is discrete but contains more unique action sequences than level $j$ has discrete actions.
    \item \textbf{Action VAE:} If the abstraction process yields continuous actions, we can make sure that the space is smooth, i.e.\ similar lower level sequences result in similar abstractions. We can achieve this through a VAE with a Gaussian distribution of the latent variable, which effectively is the abstract action mapping. Although this does not prevent model exploitation completely, it makes it less likely. Since during training of the WM, the trajectory abstraction process always yields the same results for goals, rewards and terminal flags but differs slightly for the actions, the WM learns that small variations in the abstract actions still lead to the same end result. However, large amounts of stochasticity in the abstract actions may compromise the world model performance or slow down the training. Furthermore, requiring the abstract actions to follow a Gaussian with zero mean and unit variance might be a too restrictive assumption by itself. Thus, to trade-off the benefit of local smoothness in the abstract action space with the introduced stochasticity and constraints of the VAE setting, a $\beta$ VAE 
    with a specific Gaussian prior distribution can be used. While $\beta$ allows to weigh the requirement of the abstract action distribution to follow the Gaussian prior, the variance of the prior can be set to a small value to enforce the smoothness of the abstract action space only locally.
    
\end{itemize}

\section{Experiments and Discussion}\label{sec:experimens_and_discussion}

We present a series of proof of concept experiments with a two level hierarchical model to demonstrate where the proposed MBHRL approach works. 
We explore its limitations and provide a detailed analysis of them to facilitate the search for solutions.

Our experiments presented in Section~\ref{sec:world_model_experiments} demonstrate that the hierarchical model learns meaningful and correct abstractions of the environment, which is a basic requirement for a hierarchical world model. In Sections~\ref{sec:goal_experiments} and \ref{sec:agent_performance} we furthermore show that the hierarchical agents trained via simulations from our model can solve the benchmark environments in principle. By presenting and dissecting an instance of model exploitation in the Nav2d environment in Section~\ref{sec:model_exploitation_experiments}, we provide a hypothesis of what holds back our HMBRL approach in its current form. In the same section, we build further support for our hypothesis through additional evaluations for the remaining environments, which suggest model exploitation to be at play there as well. 

\subsection{Test Environments}

\begin{figure}
    \begin{minipage}{0.5\linewidth}
        \centering
        \includegraphics[width=\linewidth]{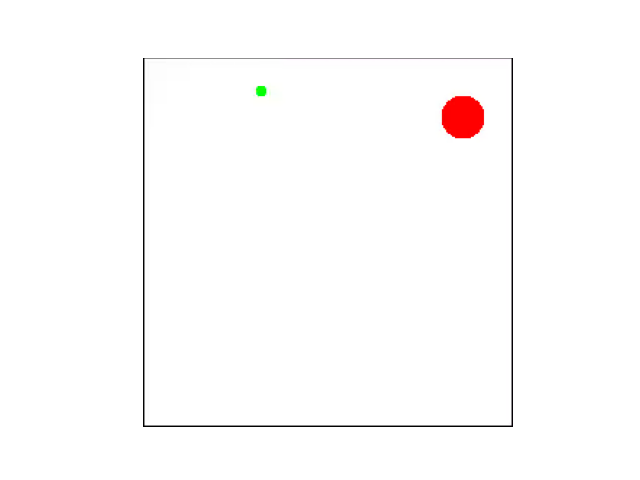}
    \end{minipage}\hfill
    \begin{minipage}{0.5\linewidth}
        \centering
        \includegraphics[width=\linewidth]{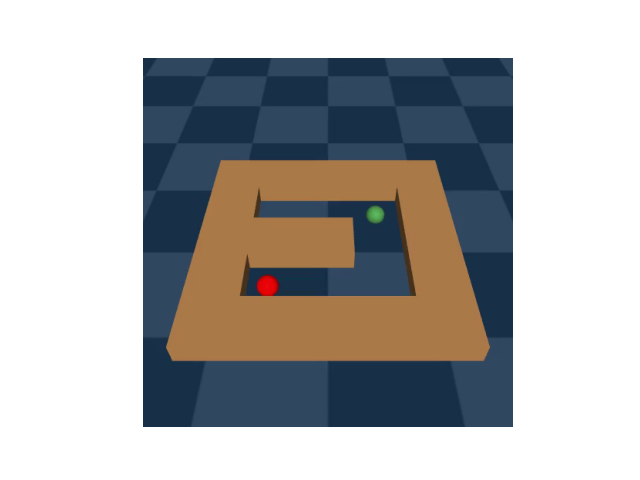}
    \end{minipage}
    \caption{\textbf{Left:} The simple ``Nav2d'' navigation environment without obstacles or momentum is based on \citet{nav2d2019github}. The agent (small green circle) can move in continuous steps across the area enclosed by the black rectangle and actions equal the displacement in x and y direction with a maximum step width limited to approximately three times the agent diameter. 
    The agent receives a negative reward of -0.01 per step until it reaches the terminal region (large red circle), which ends the episode. 
    Alternatively, the episode ends after a predefined number of steps, chosen to be 50.
    \textbf{Right:} The more complex ``PointMaze'' navigation environment features obstacles and momentum and sparse rewards. The agent (green ball) can roll around the environment in a continuous manner and is blocked by walls (orange). The actions equal the forces applied in x and y direction and the agent doesn't receive any reward unless it touches the red ball when it receives a reward of 1. 
    This environment does not terminate once the agent and the red ball come into contact, but after a predefined number of 200 environment interactions. 
    This makes it possible to collect more reward the sooner the red ball is reached by the agent.}
    \label{fig:nav_envs}
\end{figure}

\begin{figure}
    \begin{minipage}{0.5\linewidth}
        \centering
        \includegraphics[width=\linewidth]{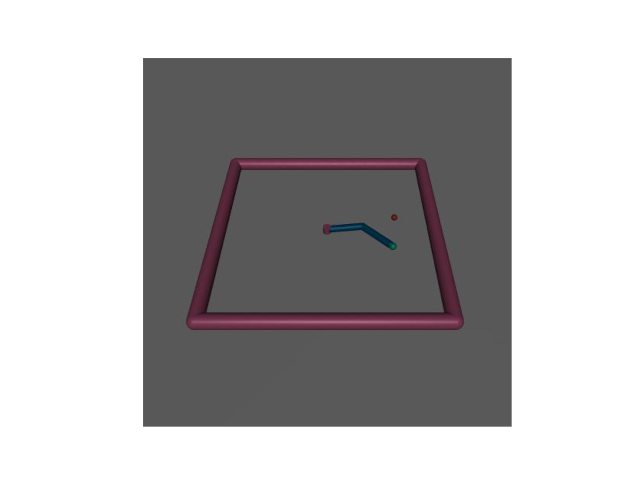}
    \end{minipage}\hfill
    \begin{minipage}{0.5\linewidth}
        \centering
        \includegraphics[width=\linewidth]{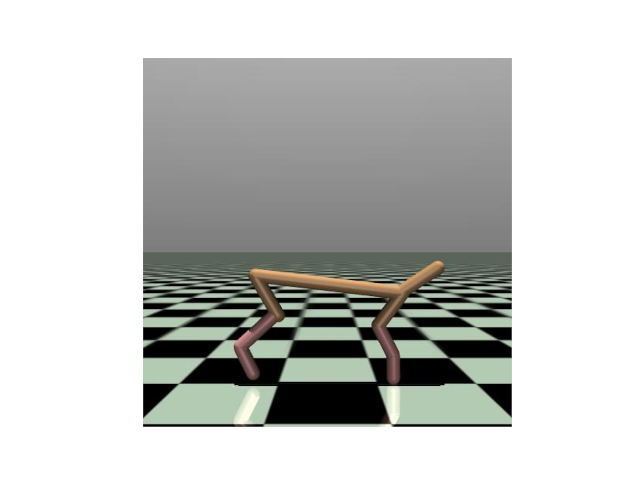}       
    \end{minipage}
    \caption{\textbf{Left:} The Reacher environment emulates a simple robotics setting where a two-jointed robot arm's end effector (green ball) has to be moved to a given position (red ball). Environment observations comprise the joint angles, angular velocities and the distance to the red ball. The arm can only be moved in two dimensions by specifying torques to its individual joints. The reward is the sum of the Euclidean distance between the end effector and the target position and a control penalty that equals the norm of the action vector. 
    \textbf{Right:} The ``HalfCheetah'' environment represents a high dimensional locomotion task with complex dynamics. Observations contain positions, velocities and angular velocities of the cheetah's joints. The task is to make the cheetah run as fast as possible to the right by applying torques to the hip, knee and ankle joints. 
    The reward depends on the running velocity minus a penalty for control cost that depends on the magnitude of the applied torques.}
    \label{fig:other_envs}
\end{figure}

To test our approach, we use the simple Nav2d navigation scenario devoid of inertia or obstacles and the more complex Pointmaze environment that includes both (see Figure~\ref{fig:nav_envs}). 
The agent starts at a random position and is tasked to reach a target positioned at another random location, associated with a reward. 
The observations contain the agent's current position as well as the target position. 
We additionally evaluate our approach on the Reacher and HalfCheetah tasks from the robotics domain described in Figure~\ref{fig:other_envs}. 
In the Reacher environment, the agent has to control a robot arm with two joints to move its end effector to a given target position on a 2d plane that is placed randomly in every episode.
Actions are in the form of torques for the individual joints and the scenario includes friction and inertia.
The HalfCheetah environment consists of a two-legged robot with 6 controllable joints that can move into two directions.
The the task is to move forward as fast as possible within a maximum amount of allowed actions.
Similar to the Reacher task, actions are provided as torques for the individual joints and the environment features inertia and friction forces. 
Both scenarios depart from the navigation domain where decision making can mainly be done based on the position data provided in the observations.
Nonetheless, Reacher's dynamics can still be considered simple as it only has two degrees of freedom and does not require sophisticated exploration. 
In contrast to that, the HalfCheetah environment represents a high dimensional, complex locomotion problem that requires precise, coordinated and well timed actions.

\subsection{World Model Accuracy}\label{sec:world_model_experiments}

We evaluated the predictive power of the hierarchical world model using exclusively trajectory data from the replay buffer. This removes the agent's actions as a possible cause for model exploitation (c.f. Section~\ref{sec:model_exploitation}) from the experimental setting and provides an isolated picture of the model's capabilities. To gain a broader picture of the model's predictive capabilities and the impact of the accumulating error that inevitably occurs in open loop rollouts, we first started a rollout closed loop for some steps, akin to a warm-up or burn-in period, and then switched to open loop at some point.

\begin{figure}[h]
    \begin{minipage}{0.01\linewidth}
        \rotatebox{90}{\small Level 0}
    \end{minipage}
    \begin{minipage}{0.99\linewidth}
        \centering
        \includegraphics[width=\linewidth]{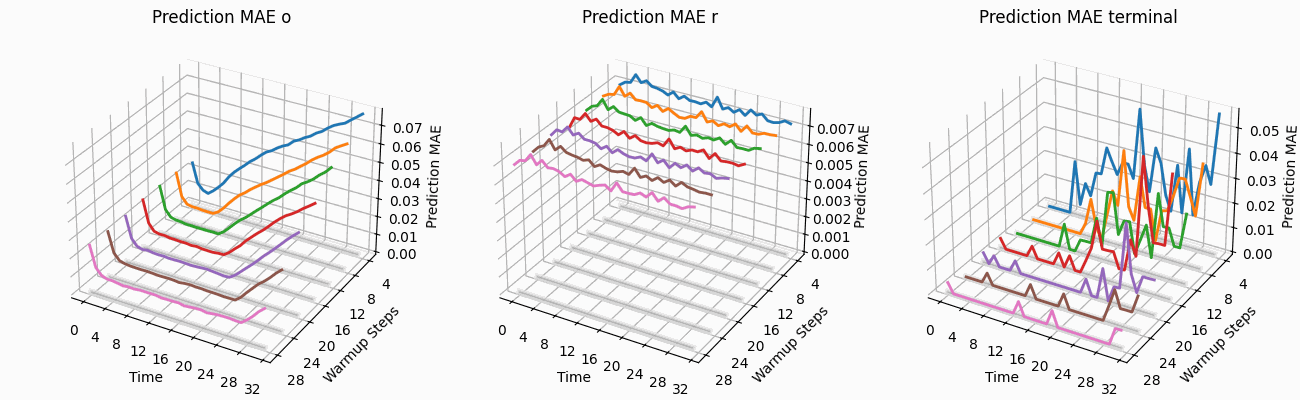}
    \end{minipage}
    \hfill
    \vspace{20px}
    \begin{minipage}{0.01\linewidth}
        \rotatebox{90}{\small Level 1}
    \end{minipage}
    \begin{minipage}{0.99\linewidth}
        \centering
        \includegraphics[width=\linewidth, trim=0 0 0 30px, clip]{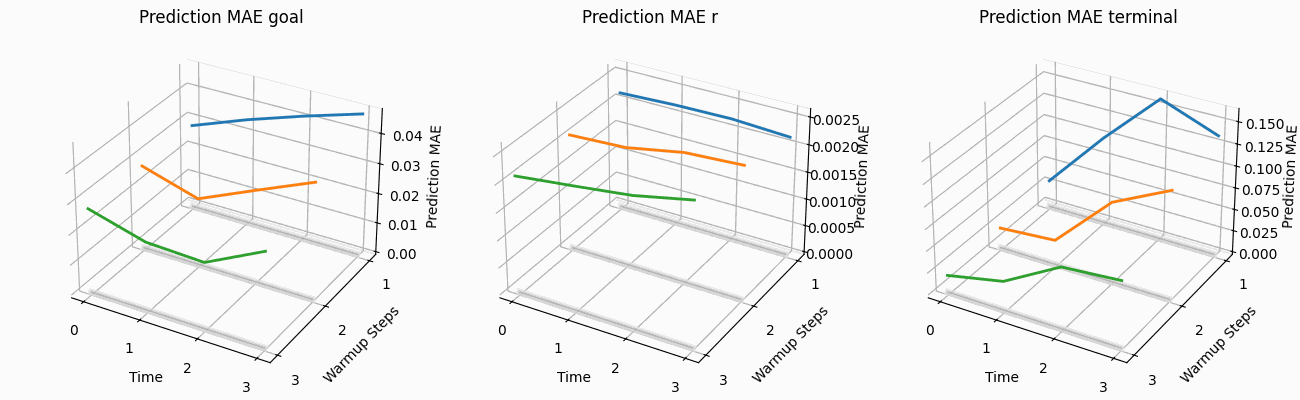}
    \end{minipage}
    \caption{Reconstruction MAE of the world model levels 0 (top row) and 1 (bottom row) in the Nav2d environment, averaged over multiple 32 step long trajectories. As the temporal stride on level 1 is 8, the total amount of model steps is 4 for there. The predicted quantities are observation/goal (left column), reward (middle column) and terminal flag (right column). The x-axis is time dimension of the trajectories, the y-axis is the amount of warm-up steps and the z-axis is the MAE. The first time step carries index 0, as there the model did just absorb the observation and reconstructed it, which is technically no prediction forward in time.}
    \label{fig:model_prediction_error}
\end{figure}

Figure~\ref{fig:model_prediction_error} shows the mean absolute error (MAE) between ground truth data and model predictions for both model levels and various amounts of warm-up followed by open loop rollout steps. 
The prediction error remains low in the warm-up period (shown on the y-axis of all plots), and steadily increases thereafter irrespective of the predicted quantity and the model level, as would be expected.
Note that the Nav2d environment has a very simple reward structure, which explains why the reward prediction error is varying around the same value for all prediction steps, both model levels and independently of the amount of warm-up steps.
We see that both levels show the highest prediction error in the observations/goals (left column), which makes sense as they are higher dimensional than rewards (middle column) and terminal flags (right column) and have a more complex structure.
It is important to differentiate between the relevance of the level 0 observation reconstruction error and the level 1 goal reconstruction error, though. 
Factoring in that reward and terminal flag errors are mild, a high observation reconstruction error can be seen as an indicator for the level 0 model struggling with the complexity of the observations. However, since the observation reconstruction is  merely done to provide a training signal for the level 0 model, a high observation reconstruction error is not a concern per se as long as the model picks up enough information from the training data to properly model the reward and terminal flag dynamics. Only they and the world model states are relevant for the agent training. A high goal reconstruction error however affects the quality of the goals that the level 0 GSA obtains. As such, the goal reconstruction error is directly related to the whole algorithm's performance, whereas the observation reconstruction error on level 0 is not.

\begin{figure}[h]
    \begin{minipage}{0.45\linewidth}
        \centering
        \includegraphics[width=\linewidth]{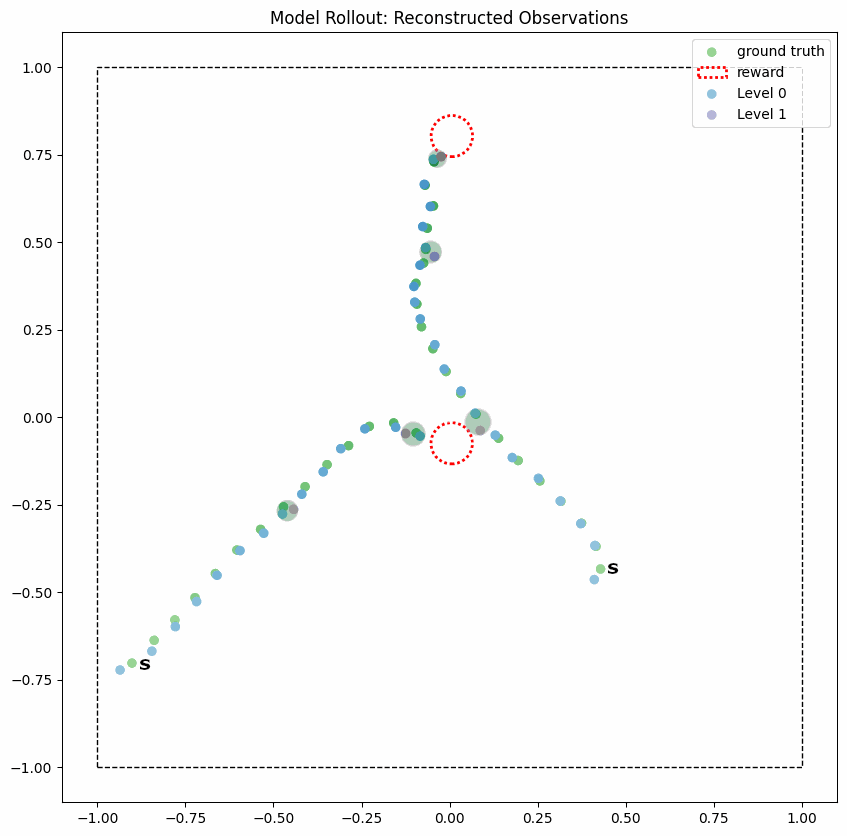}
    \end{minipage}\hfill
    \begin{minipage}{0.45\linewidth}
        \centering
        \includegraphics[width=\linewidth]{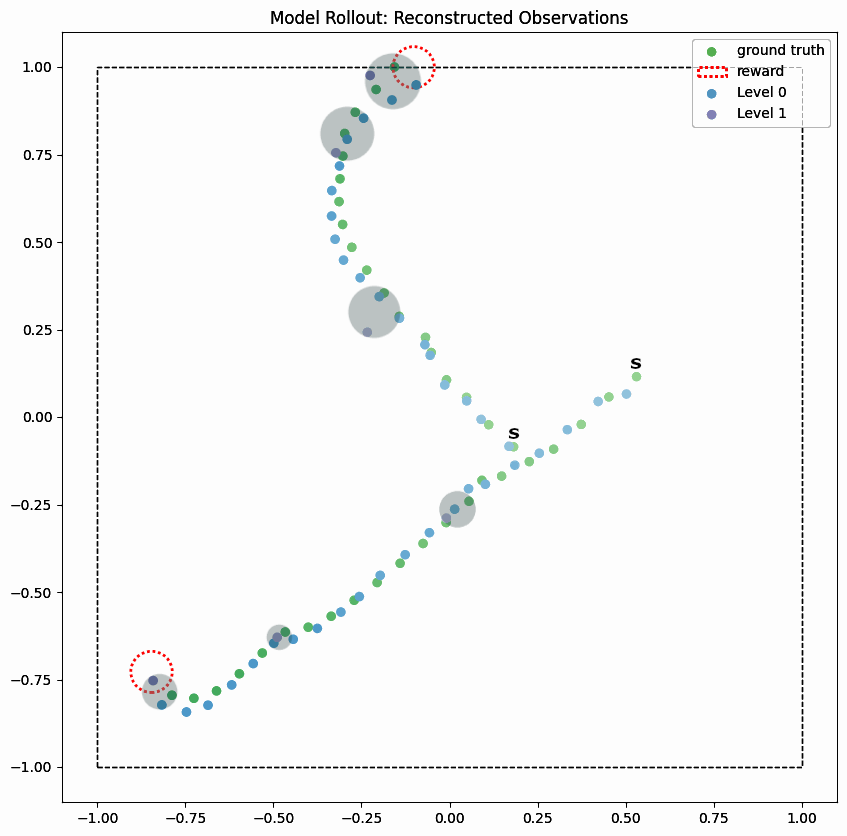}
    \end{minipage}
    \caption{Visualisation of model rollouts in the Nav2d environment. Shown are the observations reconstructed from the model states/goals via the observation decoder. Darker colour indicates later steps in time, the starting point of trajectories is marked with the letter $s$ and the terminal region is circled in red. To better illustrate the spread between ground truth and reconstructions from model rollouts, every 8th time step of the ground truth and reconstructed observations is enclosed with a shaded grey circle. The circle is as small as possible while at the same time covering the reconstructed observations from both levels as well as the ground truth. Larger circles highlight a bigger spread between the observations, suggesting a less precise reconstruction.
    \textbf{Left:} Two closed loop rollouts as a baseline for comparison. While decline in accuracy compared to the ground truth observations is visible, the reconstructed observations unsurprisingly stay more on track, particularly for the level 1 model.}
    \textbf{Right:} Two open loop rollouts with 3 and 1 warm-up step(s) for level 0 and 1, respectively. Indicating mildly accumulating model error, the reconstructed observations show a larger spread compared to the closed loop rollouts. Furthermore, the effects of noise and accumulating error are more pronounced for the observations reconstructed from the level 1 model.  
    \label{fig:rollout_observation_reconstruction}
\end{figure}

To understand if the goals proposed by the level 1 model in our experiments are usable for the level 0 GSA, we investigated the level 0 observations that can be in reconstructed from the goals. Four representative example reconstructions are shown in Figure~\ref{fig:rollout_observation_reconstruction}, two closed loop rollouts (right) and two with a warm-up period of 3 and 1 steps for the level 0 and level 1 models respectively (left). The open loop rollout experiments emulate the agent training as there we take starting points from real trajectories and perform short open loop rollouts using agent actions from there on. The accumulating prediction error known to occur in open loop rollouts \citep{holland2019effect, janner2021trust} manifests through higher deviation between reconstructed and ground truth observations towards the end of the trajectories. While the closed loop rollouts are more precise, this is not by a large margin and the reconstruction error for both model levels behaves within expectation considering other works that use the RSSM \citep{hafner2020dream, hafner2022deep}.
The combined numerical evaluation of the prediction error and manual inspections as illustrated in Figure~\ref{fig:rollout_observation_reconstruction} suggest the world model accuracy to be tolerable for the intended usage scenario. 

\subsection{Goal Representation and Similarity Measure}\label{sec:goal_experiments}

Similar to \citet{nachum2018data}, in the navigation environments and Reacher we found comparing reconstructed observations to be superior to comparing world model states directly. While using the RSSM states did work as well, it resulted in slightly slower learning. However, in the HalfCheetah environment we obtained the best results by using the world model states. We conjecture that the more complex dynamics of the environment and the higher prediction error of the world models rendered the observations too imprecise to be usable for goal navigation.

\begin{figure}[h]
    \centering
    \includegraphics[width=\linewidth]{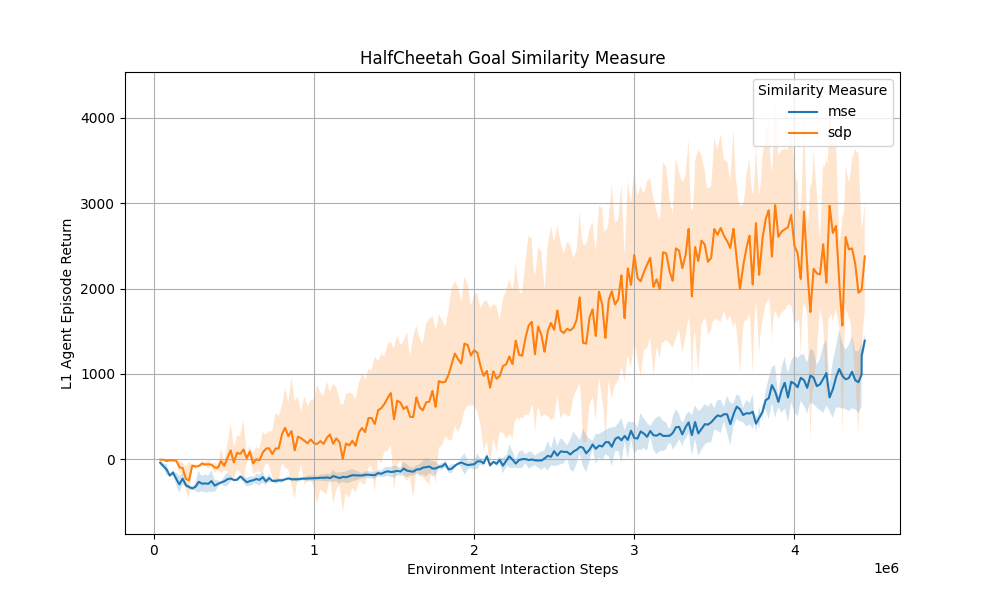}
    \caption{Comparison of the goal similarity measures $\mathcal{M}_{MSE}$ and $\mathcal{M}_{SDP}$ used in our hierarchical MBRL approach. The results were obtained by averaging three independent runs per similarity measure with a temporal stride of four steps for the level 1 model. The shaded regions indicate the standard deviation and the abstract action dimension was 16 in all runs.}
    \label{fig:goal_similarity_measure_comparison}
\end{figure}

We found the goal similarity measure to have little impact on the navigation environments and Reacher. Our assumption is that those environments are simple enough so that different goal similarity measures have no opportunity to affect the learning process measurably. However, this was not the case for the HalfCheetah environment. To show the impact of the goal similarity measure there, we conducted three training runs using the $\mathcal{M}_{MSE}$ from Equation~\ref{eqn:goal_similarity_mse} and three runs using $\mathcal{M}_{SDP}$ from Equation~\ref{eqn:goal_similarity_ndp} and averaged their results per group. The average return plotted against the amount of environment interaction steps is shown in Figure~\ref{fig:goal_similarity_measure_comparison}. While the max cosine similarity yielded higher maximum performance, it did result in much higher variance trajectories and sometimes a complete breakdown of performance below chance level. We conjecture that the max operator for the normalisation in Equation~\ref{eqn:goal_similarity_ndp} may introduce sudden and unexpected jumps in $\mathcal{M}_{SDP}$ that can destabilise the agent training. In contrast to that, the mean squared error produced slower training progress and lower asymptotic performance but much higher training stability, which is why we chose it for the Mujoco experiments presented in Section~\ref{sec:agent_performance}. 

\subsection{Agent Performance}\label{sec:agent_performance}

\begin{figure}
    \begin{minipage}{0.45\linewidth}
        \centering
        \includegraphics[width=\linewidth]{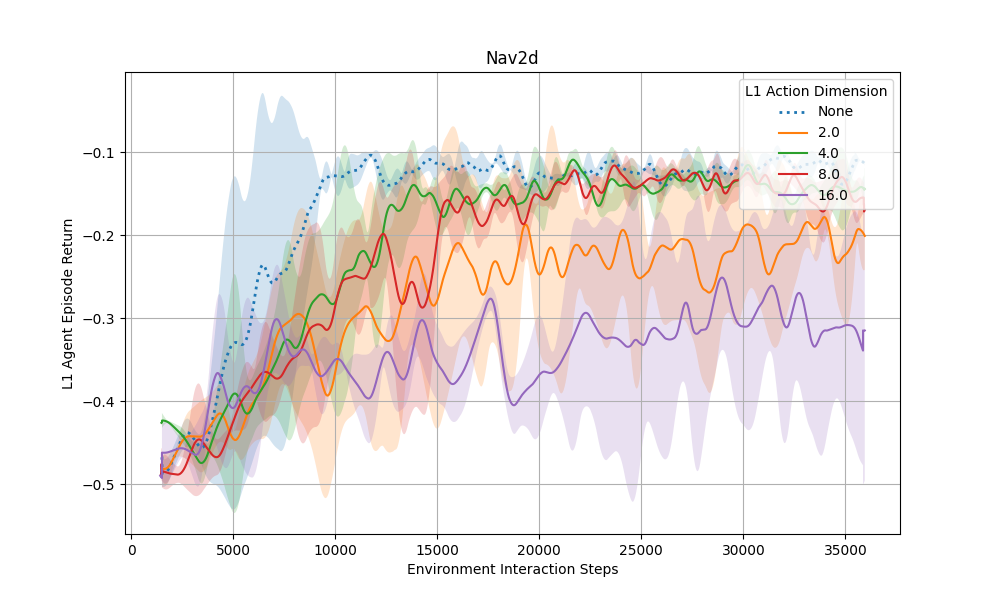}
    \end{minipage}\hfill
    \begin{minipage}{0.45\linewidth}
        \centering
        \includegraphics[width=\linewidth]{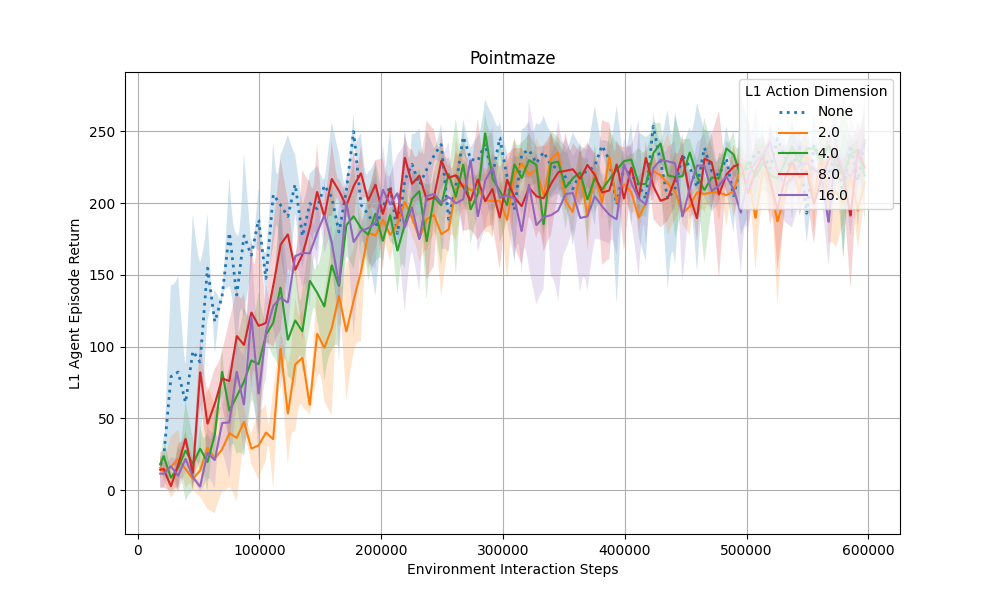}
    \end{minipage}
    \caption{Average episode return as a function of the number of environment interactions. We tested a non-hierarchical (blue, dotted line) MBRL baseline and a two level hierarchical (red, green, orange, purple solid lines) MBRL agent with varying action dimension of the level 1 world model. Each line is the average of three independent runs, shaded areas are the standard deviation.
    \textbf{Left:} Episode return in the Nav2d environment. While the non-hierarchical baseline initially learns faster, the hierarchical approach needs more time presumably to tune its individual levels to the environment. A clear sweet spot of 4 or 8 for the abstract action dimension is visible.
    \textbf{Right:} Episode return in the point maze environment. In this environment, the hierarchical approach performs very similar to the baseline with a tendency of higher abstract action dimensions to learn faster.}
    \label{fig:experiment_episode_return_hierarchical_navigation}
\end{figure}

\begin{figure}
    \begin{minipage}{0.45\linewidth}
        \centering
        \includegraphics[width=\linewidth]{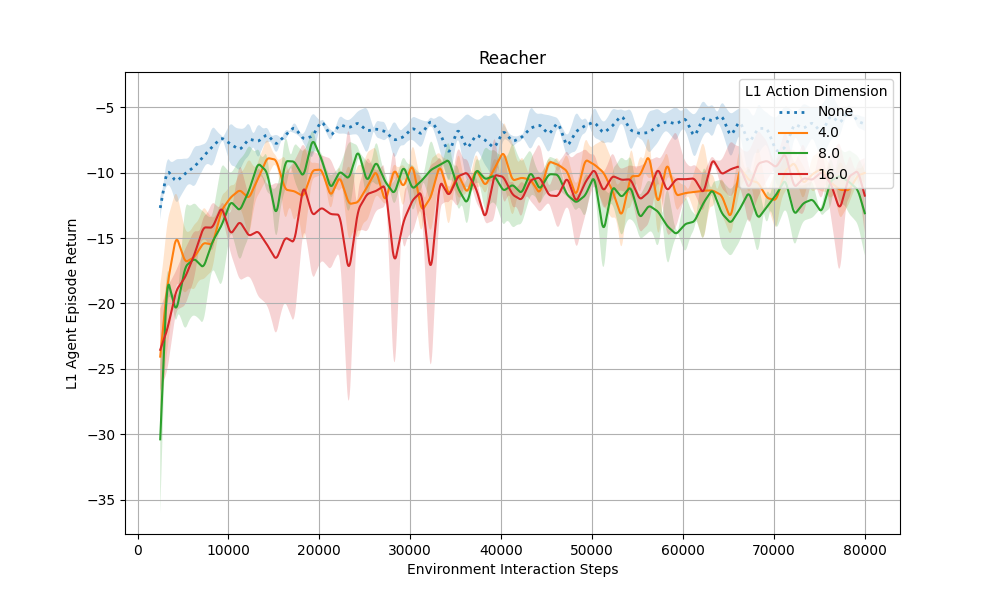}
    \end{minipage}\hfill
    \begin{minipage}{0.45\linewidth}
        \centering
        \includegraphics[width=\linewidth]{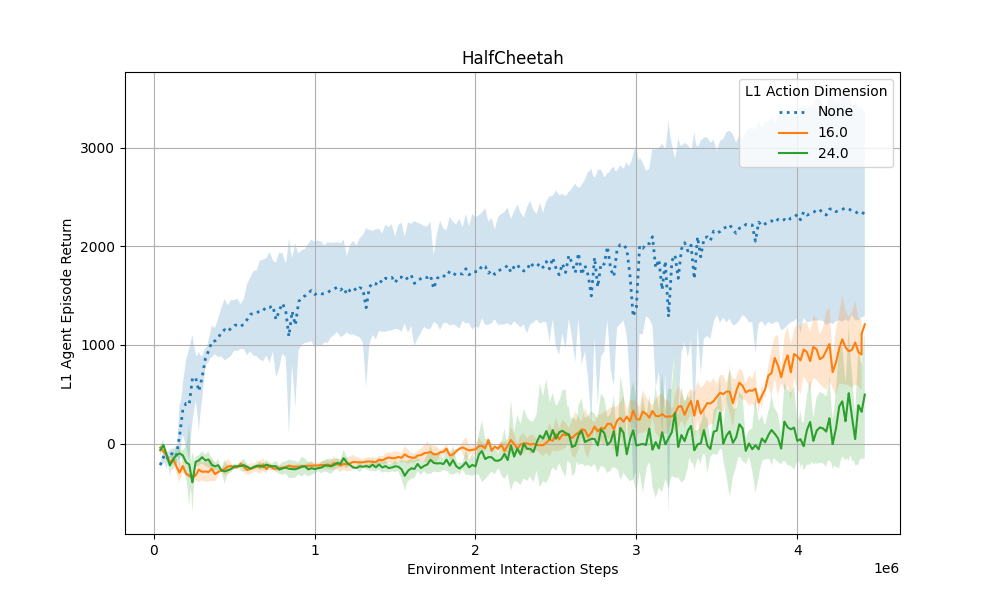}
    \end{minipage}
    \caption{Average episode return as a function of the number of environment interactions. We tested a non-hierarchical (blue, dotted line) MBRL baseline and a two level hierarchical (red, green, orange solid lines) MBRL agent with varying action dimension of the level 1 world model. Each line is the average of three independent runs, shaded areas are the standard deviation.
    \textbf{Left:} Episode return in the Reacher environment. The performance deficit to the non-hierarchical baseline remains throughout the training and a mild influence of the abstract action dimension is visible.
    \textbf{Right:} Episode return in the HalfCheetah environment. The performance gap of the hierarchcial approach to the baseline is more pronounced than in the other environments. The abstract action dimension has a noticeable impact on performance, yielding stronger results in case of 16-dimensional abstract actions compared to 24-dimensional abstract actions.}
    \label{fig:experiment_episode_return_hierarchical_robotics}
\end{figure}

An advantage of our HMBRL approach is the action abstraction process that leads to comparatively low dimensional abstract actions. To shed light on the impact of different abstract action dimensions, we evaluated the final episode return of our approach on the test environments with varying abstract action dimensions as shown in Figures~\ref{fig:experiment_episode_return_hierarchical_navigation} and \ref{fig:experiment_episode_return_hierarchical_robotics}. For that, we ran each configuration with three independent random seeds and averaged the results obtain the presented graphs. Across all environments, our approach performs at best on par with a non-hierarchical baseline version that is essentially the world model presented in \citet{hafner2020dream}. While we discuss possible reasons unique to the individual environments in the following, we found model exploitation to play a role in the presented environments as well. This aspect is discussed in more detail in Section~\ref{sec:model_exploitation_experiments}.

For the Nav2d environment (Figure~\ref{fig:experiment_episode_return_hierarchical_navigation} left) we used a temporal stride of 8 time steps on level 1. While the non-hierarchical baseline is slightly faster than our approach in terms of learning speed, it is on par for final performance for two of the four tested abstract action dimensions. The initial performance deficit of the hierarchical model compared to the non-hierarchical baseline is likely due to the additional model components that have to be trained and need time to converge on the environment dynamics. The performance of the hierarchical model peaks when using abstract action dimensions of 4 or 8, which is a reduction of 75\% or 50\% percent compared to the 16-dimensional sequence of 8 ground truth actions. If the abstract action dimension is chosen too high, the model performance declines. We conjecture that a higher-dimensional action space is harder to explore due to the curse of dimensionality and encourages the occurrence of abstract action sequences that are unfamiliar to the model (c.f. Section~\ref{sec:model_exploitation_experiments}). Note that 

In the more complex Pointmaze environment (Figure~\ref{fig:experiment_episode_return_hierarchical_navigation} right), the temporal stride from the Nav2d environment is maintained. The hierarchical and non-hierarchical model exhibit very similar performance, although the latter learns faster during early training stages. Again, a reasonable explanation for this is the need for training of additional components in the more complex hierarchical world model. We assume that the very similar performance of the hierarchical and the non-hierarchical models is caused by the movement momentum in the environment. It smoothes out the consequences of actions, making far-sighted acting more and precise actions less relevant. We conjecture that this acts as a regularising effect to combat model exploitation and offsets the additional cost of having to train the additional components of the hierarchical model to some degree . The level 1 action dimension has a less noticeable effect here compared to the Nav2d environment. Yet, 8 and 16 dimensions seem to work equally well. While an 8-dimensional abstract action versus a sequence of 8 2-dimensional level 0 actions is not a compression as strong as in the Nav2d environment, it still has only half as many dimensions.

In the Reacher environment (Figure~\ref{fig:experiment_episode_return_hierarchical_robotics} left), we again use a temporal stride of 8 on level 1. The hierarchical approach maintains a small but constant performance gap to the baseline. While a 4-dimensional abstract action dimension seems to be initially inferior to higher dimensional abstract actions, the final performance is best among all configurations. Similar to the Pointmaze environment, movement momentum is a strong candidate for explaining the similar performance of the various abstract action dimensions.

The performance difference of our hierarchical MBRL approach to the baseline in the HalfCheetah environment (Figure~\ref{fig:experiment_episode_return_hierarchical_robotics} right) is more pronounced than in all other tested environments. We conjecture that, independently of the model exploitation issues, the reactive and timing sensitive nature of the tasks is not well suited for a hierarchical approach where higher level decision making reacts by design increasingly slower. Manual inspection of interaction videos for this environment support our hypothesis, as there the hierarchical agent was able to learn movement patterns for the individual legs but often seemed incapable of properly synchronising them to produce a reliable forward motion. As a result, in those situations the cheetah performed forward-backward motions that more or less cancelled each other out and diminished the overall return. 

\subsection{Model Exploitation Experiments}\label{sec:model_exploitation_experiments}

\begin{figure}[h]
    \centering
    \includegraphics[width=\linewidth]{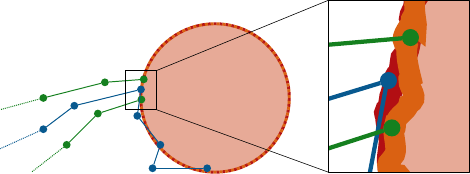}
    \caption{A model exploitation scenario in the Nav2d environment provoked by the level 1 agent, visualised via level 0 observations.
    \textbf{Left:} Three trajectories leading to the terminal region of the Nav2d environment. The large red-orange circle marks the terminal region, where the agent obtains at the same time a reward of 1. The two green trajectories are generated from temporally abstracted ground truth data, the blue trajectory is an open loop rollout of the level 1 RMA. It shows the jittering behaviour found by manual inspection of the agent environment interaction and the level 1 RMA training.
    \textbf{Right:} The border of the terminal region up-close. The level 1 world model is not completely accurate here outside of the regions trained via ground truth data. The reward border (red line) at which the model starts to predict positive reward is not everywhere perfectly aligned with the termination border (orange) at which the model starts to predict a high episode termination probability. This mismatch between the borders of the two regions, which should actually be perfectly superimposed, can be expected from the world model as it is not trained with ground truth data everywhere. At the same time, it is a reasonable assumption that over time the level 1 RMA finds these inaccuracies and exploits them for high reward.}
    \label{fig:model_exploitation}
\end{figure}

As discussed in Section~\ref{sec:model_exploitation}, model exploitation can occur in the hierarchical levels of the world model when using the action autoencoder. To investigate this issue more closely, we designed an experiment in a modified Nav2d environment where the agent receives a reward of 1 when entering the terminal region instead of not being rewarded. In comparison to not rewarding the agent, the hierarchical approach struggles notably more in this setting. By manual inspection of the collected data, we noticed a particular behaviour pattern: The hierarchical approach learned to move towards the terminal region in a very direct and efficient manner, only to jitter closely around it for a small amount of steps or to move away again. An investigation of the simulated model rewards showed that the level 1 agent sometimes achieved positive rewards in multiple time steps while at the same time not triggering a terminal flag of 1, which is not possible in the real environment. Inspecting the simulated training trajectories of the level 1 agent in detail, we came across the same jittering pattern first found in the collected environment data. Since we already learned from the model experiments that the level 1 model predicts temporally abstracted real world trajectories with reasonable accuracy, our hypothesis is that the level 1 agent seeks and exploits inaccuracies of the level 1 model at the border of the terminal region where it is not directly grounded by real data. An explanatory visualisation of our hypothesis can be found in Figure~\ref{fig:model_exploitation}.

\begin{figure}[ht]
    \centering
    \includegraphics[width=\linewidth]{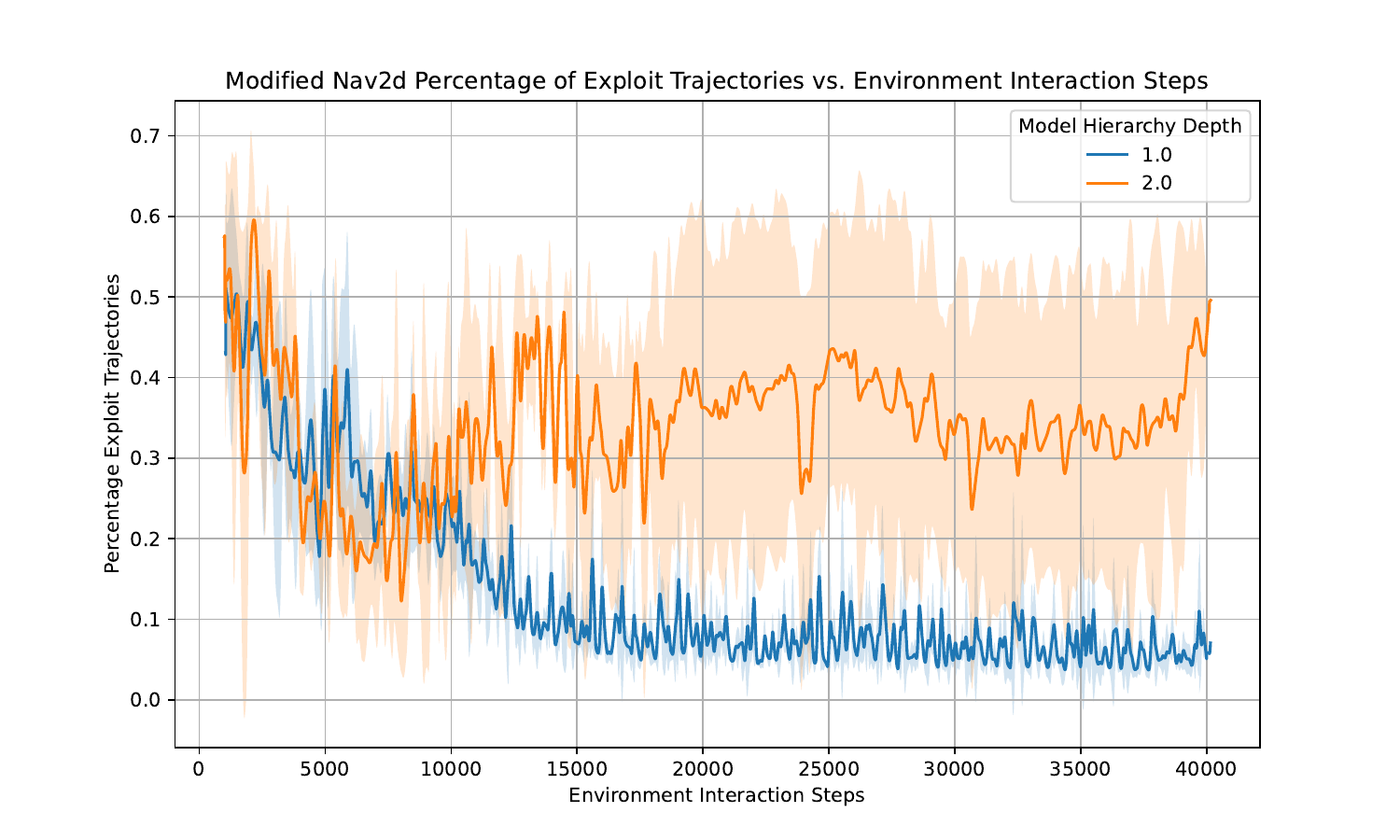}
    \caption{The fraction of agent training trajectories with more than one step of positive reward vs. the amount of collected environment data in the modified Nav2d environment. The level 1 agent of the hierarchical model (orange) produces notably more unrealistic trajectories with more than one step of positive reward, hinting at a model exploitation scenario. Compared to that, the level 0 agent on the non-hierarchical model produces increasingly less exploitative trajectories over the course of training.}
    \label{fig:perc_exploit_agent_trajectories}
\end{figure}

Over time, the level 1 agent learns to navigate to those regions of the terminal zone border where the reward is already above zero but the model does not yet predict the episode to terminate. To test this hypothesis, we compared the average fraction of open loop agent training trajectories with more than one step of positive reward for a hierarchical and a non-hierarchical agent. The result can be seen in Figure~\ref{fig:perc_exploit_agent_trajectories}, the hierarchical approach produces substantially more trajectories that have more than one step of rewards, which we interpret as exploitative trajectories. While being not entirely absent from the non-hierarchical agent training either, it can be seen that the total fraction of exploitative trajectories continually falls over the course of training and quickly reaches a stable, low value at around 15000 to 20000 environment interaction steps. Intriguingly, this is shortly after the epsilon-greedy exploration phase of the agent ends. Since at this point the only exploration incentive comes from the entropy term in the agent loss, the agent performs considerably fewer untypical actions overall. Inevitably, through real world data the majority of model inaccuracies have been repaired by now, which is indeed possible for the level 0 model. Since the agent does not explore as strongly anymore, only a small amount of new ones are found and exploited, as we can see with the blue graph in Figure~\ref{fig:perc_exploit_agent_trajectories}.

\begin{figure}[h!]
    \centering
    \includegraphics[width=\linewidth]{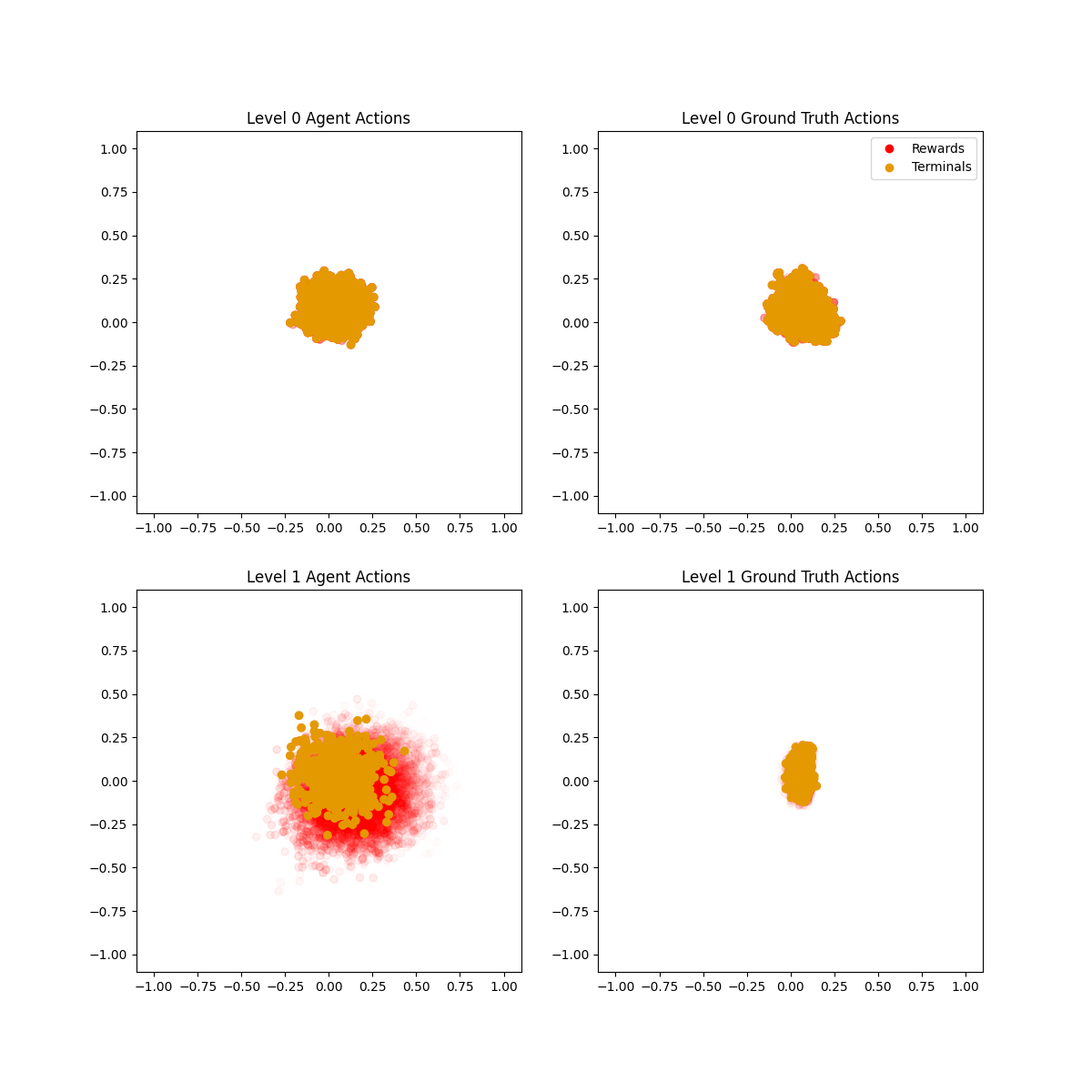}
    \caption{Plots of open loop rollouts from a non-hierarchical baseline model in the top row and the level 1 model of our hierarchical approach in the bottom row. To visualise the reward and terminal flag dynamics of the environment model, only simulation steps with a positive reward or terminal flag are plotted in red and orange, respectively. The position of the markers is determined through the predicted observation from the open loop rollouts. Their opacity depends on the particular value, whereas a terminal flag or reward of 0 results in fully transparent and a value of 1 in fully opaque markers. Rewards below 0 or above 1 have been clamped for simplicity. While for the level 0 rollouts the terminal and reward regions correctly overlap, for the level 1 rollouts this is only the case if actions obtained through temporal abstraction were used. The rollouts guided by the level 1 RMA's actions result in unrealistic and wrong simulations that permit to collect reward without having to end an episode.}
    \label{fig:model_exploitation_nav2d}
\end{figure}

To further support our hypothesis, we devised an experiment to recreate the situation depicted in Figure~\ref{fig:model_exploitation} from empirical data. In this experiment, open loop model rollouts of the modified Nav2d environment were performed either with ground truth actions from the replay memory or actions proposed by the RMA. For data collection, the agent starting positions were randomly sampled from a ring shaped area around the terminal region and actions were selected using a handcrafted optimal policy. The resulting trajectories shared the same average length and approached the terminal region from all possible angles, thereby providing dense coverage of the environment's state space. With a warm-up period of 4 steps for the level 0 model and 1 step for the level 1 model, open loop rollouts either using the remaining ground truth actions or those proposed by the RMA were generated. The simulated trajectories were evaluated and for each episode, the reward and episode termination probability per step were plotted using the 
predicted environment positions as shown in Figure~\ref{fig:model_exploitation_nav2d}. For the level 1 model, this experiment uncovered a severe mismatch between the regions where the reward starts to become positive and the episode termination probability raises towards 1 if the RMA's action are used to generate open loop rollouts. It is in line with the situation portraied in Figure~\ref{fig:model_exploitation} and further supports our model exploitation hypothesis.

It should be noted that although in principle the level 1 RMA can attempt model exploitation for any environment, it is not equally harmful for all of them. The unmodified Nav2d environment for example features only negative per-step rewards and navigating to the terminal region merely interrupts this influx of mild punishment. As there is no additional positive reward to gain by stepping into the terminal region, no exploitation scenario emerges. In the Pointmaze environment where the agent is in fact positively rewarded once it reaches the red ball, the episode does not end upon reaching it. This removes the necessity to find unrealistic and exploitative actions, as the environment already permits to collect many rewards in succession by staying close to the red ball and no conflict with the terminal flag exists. Even if we modify the Pointmaze enviroment to end upon reward collection, exploitation is difficult due to the momentum in the ball movement, which makes it hard to at the same time move quickly towards the red ball and still stop in time to profit from the exploitable region closely around the red ball. 

\begin{figure}[h!]
    \centering

    \begin{subfigure}[b]{0.24\textwidth}
        \centering
        \includegraphics[width=\textwidth]{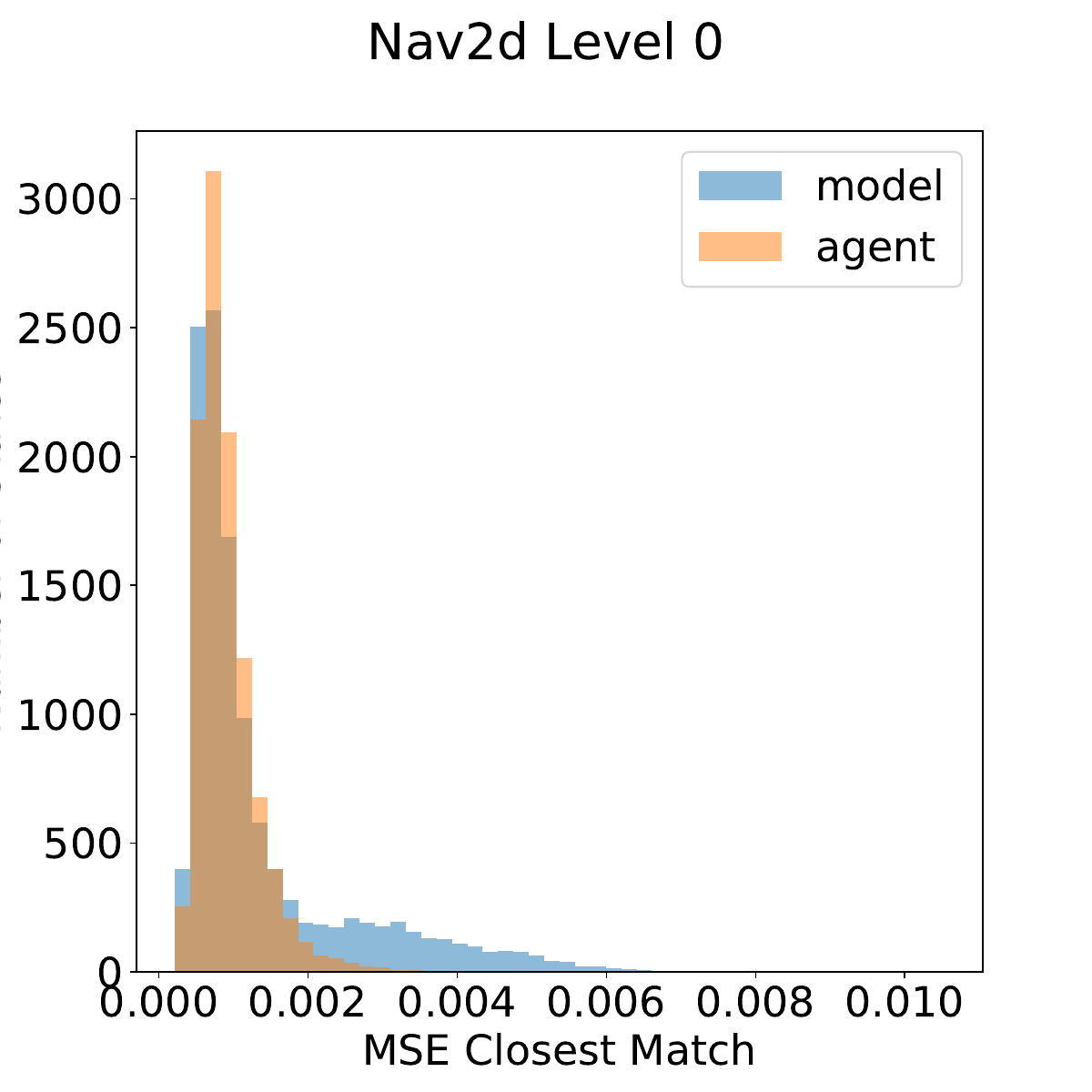}
    \end{subfigure}
    \hfill 
    \begin{subfigure}[b]{0.24\textwidth}
        \centering
        \includegraphics[width=\textwidth]{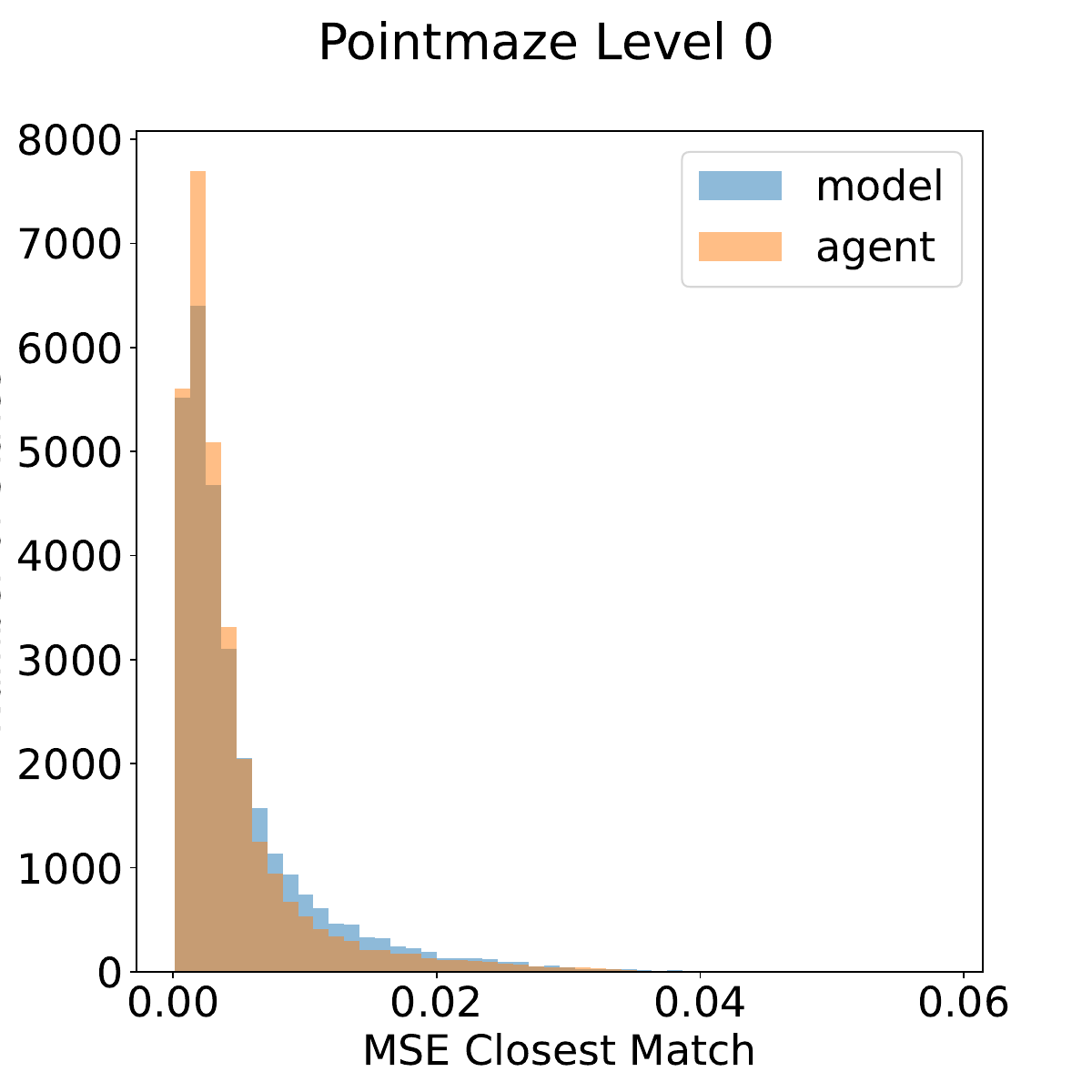}
    \end{subfigure}
    \hfill 
    \begin{subfigure}[b]{0.24\textwidth}
        \centering
        \includegraphics[width=\textwidth]{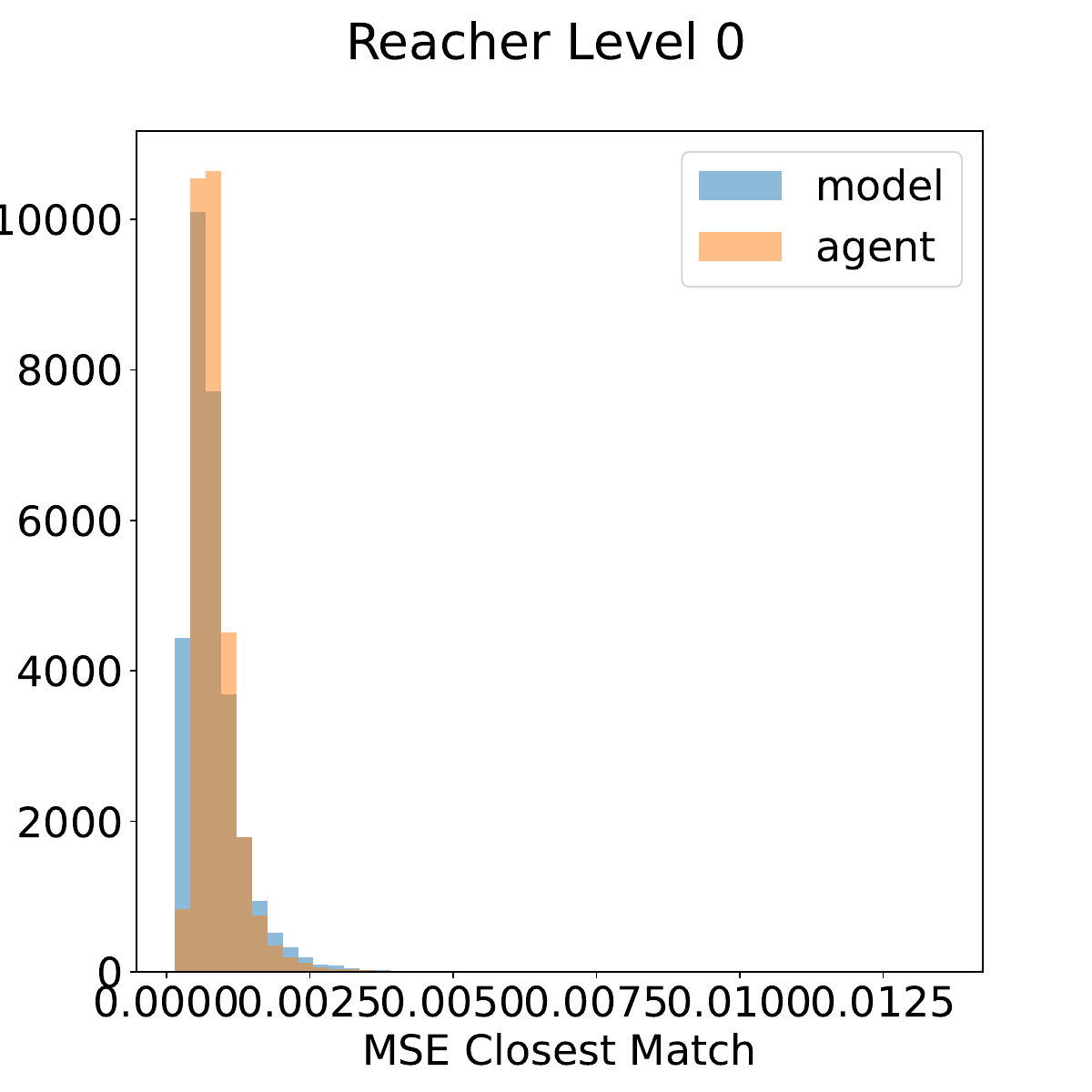}
    \end{subfigure}
    \hfill 
    \begin{subfigure}[b]{0.24\textwidth}
        \centering
        \includegraphics[width=\textwidth]{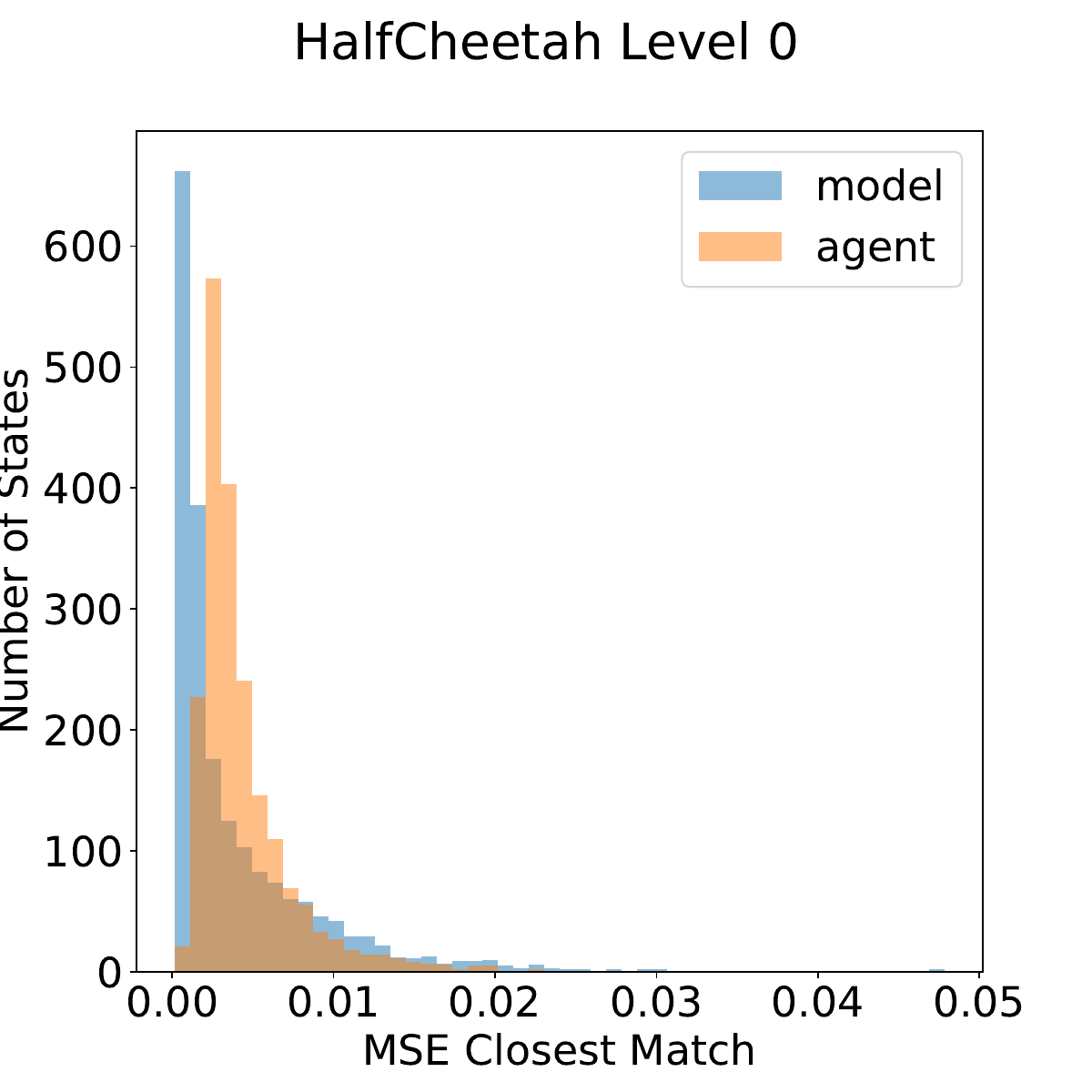}
    \end{subfigure}

    \begin{subfigure}[b]{0.24\textwidth}
        \centering
        \includegraphics[width=\textwidth]{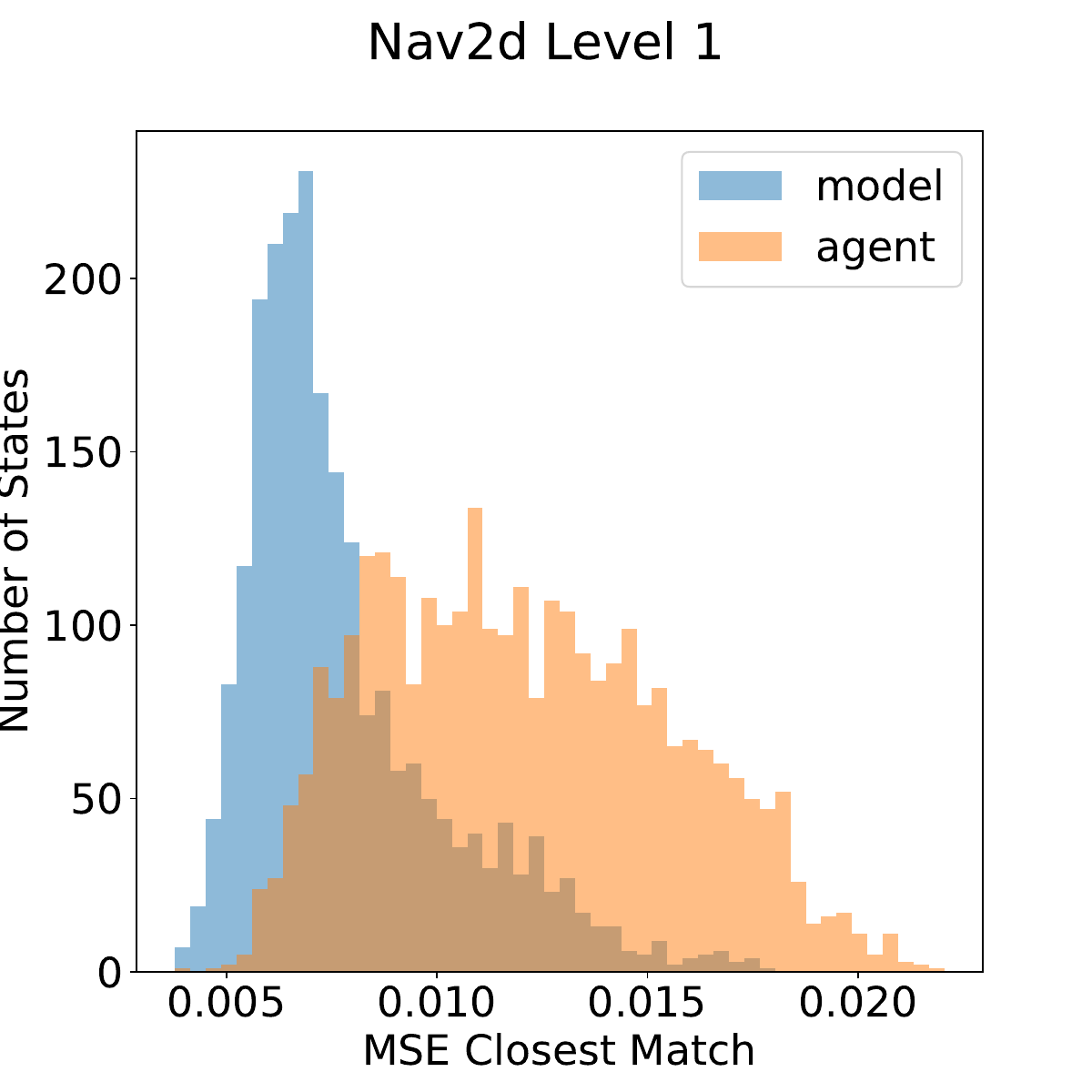}
    \end{subfigure}
    \hfill 
    \begin{subfigure}[b]{0.24\textwidth}
        \centering
        \includegraphics[width=\textwidth]{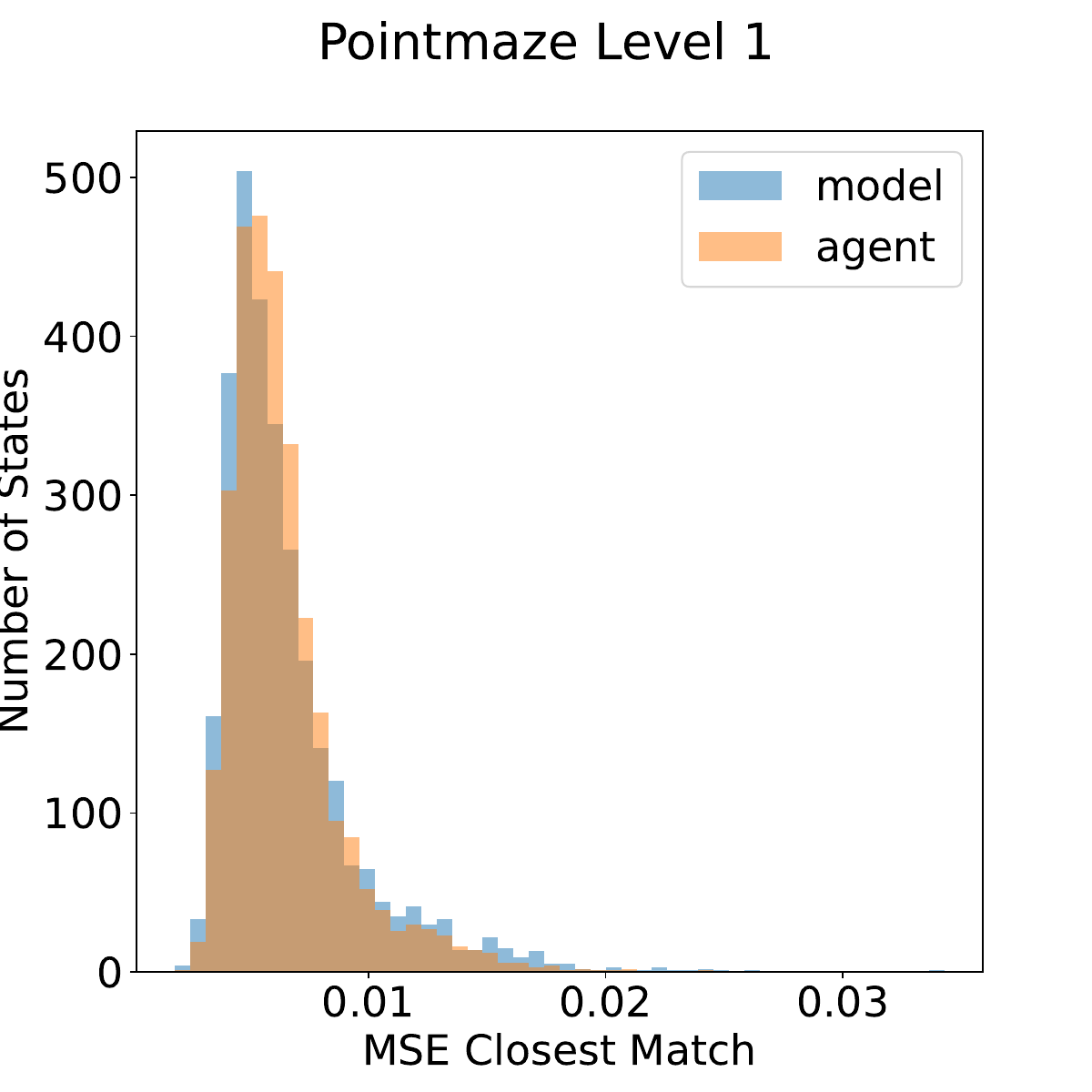}
    \end{subfigure}
    \hfill 
    \begin{subfigure}[b]{0.24\textwidth}
        \centering
        \includegraphics[width=\textwidth]{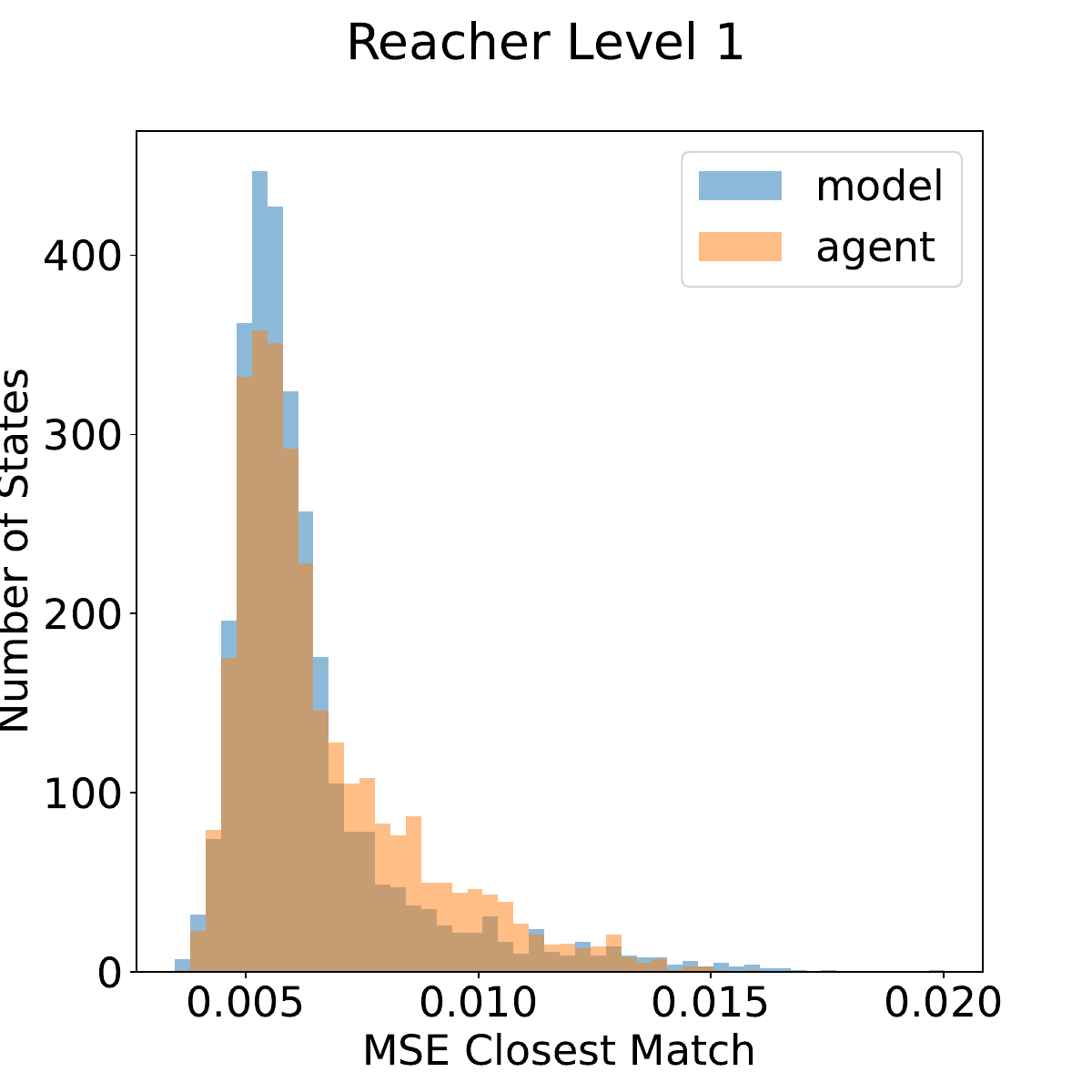}
    \end{subfigure}
    \hfill 
    \begin{subfigure}[b]{0.24\textwidth}
        \centering
        \includegraphics[width=\textwidth]{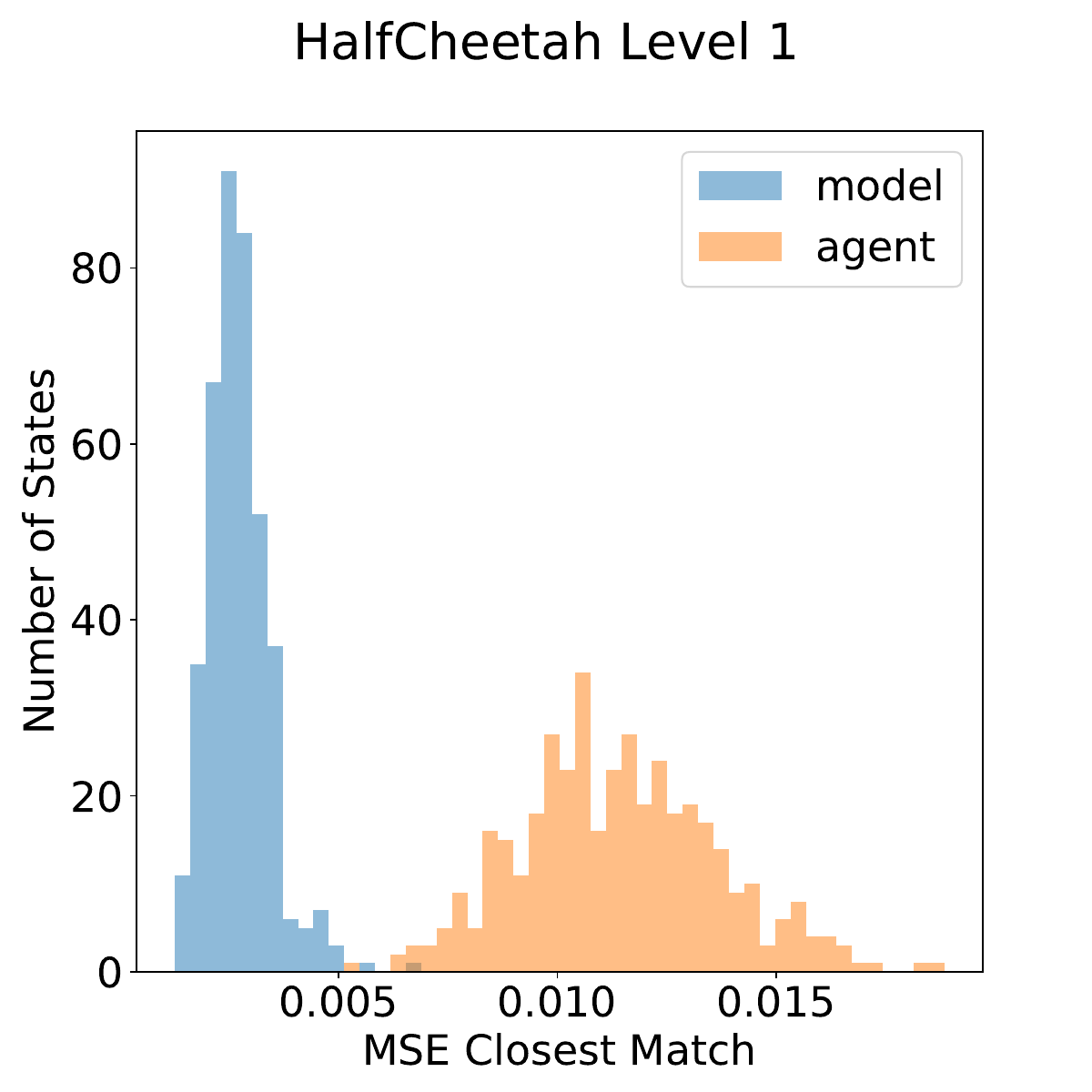}
    \end{subfigure}
    
    \caption{The histograms show the distribution of rollout state distances to the closest from a group of representative states for all four environments. Results from the Nav2d, Pointmaze, Reacher and HalfCheetah environments are depicted from left to right. The top row compares open loop rollouts using a non-hierarchical baseline model and the bottom row does the same for level 1 of a hierarchical model. Ground truth action rollouts are in blue color and agent rollouts are orange. In the first row, ground truth and agent rollouts show largely the same degree of deviation from the set of representative states. This deviation can be explained by simple compounding model errors, particularly since it occurs to the same degree in the agent as well as the ground truth action rollouts. Only the HalfCheetah results exhibit a different trend: The agent rollouts are notably more different from the representative states than the ground truth action rollouts. As HalfCheetah is the most difficult environment with the highest action dimension, we conjecture that model exploitation is occuring even in the level 0 model to a mild degree. In the second row, larger deviations for the agent rollouts indicate that the model is pushed to increasingly uncommon states due to the actions of the agent. While this deviation is not necessarily a sign of model exploitation, it is a prerequisite for it.}
    \label{fig:model_exploitation_histograms}
\end{figure}

Unfortunately, due to their more complex and less relatable reward functions, it is not as straightforward to make the same arguments for the Reacher and HalfCheetah environments. While we strongly assume that model exploitation happens here as well and see the performance gap between the hierarchical model and the baseline as supporting our assumption, we can't easily produce visualisations like Figure~\ref{fig:model_exploitation_nav2d}. To investigate whether the level 1 RMA actions do at least lead to uncommon model states in open loop rollouts, which would be a prerequisite for model exploitation, we performed a final experiment. For each environment, using closed loop rollouts we generated a library of common and familiar world model states from a blend of random trajectories and those collected by the agent. We then performed open-loop rollouts using either actions from the replay memory (temporally abstracted actions for level 1 rollouts) or agent actions, whereas in both cases the rollouts were initialised with the same warm-up steps. We then compared the open loop model and agent rollouts to the previously generated state library and recorded for each open loop rollout state the Euclidean distance to the closest match in the library. The resulting histograms are shown in Figure~\ref{fig:model_exploitation_histograms} and clearly support the assumption that the level 1 RMA actions lead to notably uncommon world model states for all environments. As outlined above, we empirically found that this does not necessarily lead to suboptimal policies depending on the reward dynamics. However, it illustrates the problematic lack of grounding that is present for the RMAs in abstract levels.

\section{Related Work}

An integral part of both HRL as well as HMBRL is the implementation of the hierarchical decision making process. Based on this, approaches in the field can be loosely divided into two groups that follow related but slightly different concepts.

One group is goal-conditioned H(MB)RL that assumes levels in the hierarchy to propose goals to their lower levels that should be achieved within a fixed amount of steps. The goals are formulated in terms of an environment state, world model state, or observation and do not necessarily carry a significant semantic meaning in the context of the hierarchy. These approaches circumvent the need for deciding when to transfer control between the levels. However, a limitation introduced by hierarchical abstraction through fixed step sizes is that some tasks might not be easily compatible with the chosen step size or inherently require a dynamic step size.

In contrast to that, approaches that build semantic hierarchies implement more complex, variable length decision making on the individual levels in the sense of the options framework \citep{sutton1999between}. They need mechanisms to decide when to transfer control to a lower level, e.g.\ if a new sub-task starts, and when to yield control back to an upper level. This introduces additional flexibility at the cost of complexity that might nonetheless be necessary to succeed in certain environments.

\subsection{Goal-Conditioned H(MB)RL}


Conceptually heavily inspired by \citet{dayan1993feudal}, \citet{vezhnevets2017feudal} use a high level manager policy that proposes via its actions goals to a low level worker policy. Both policies use an embedding network to preprocess observations and the goals given by the manager are desired embedded observations.
\citet{vezhnevets2017feudal} find several crucial components to make the hierarchical setting work. Directional, relative instead of absolute goals allow for better generalisation of the worker. A special training method of the manager replaces during manager training actual subtrajectories produced by the worker with a statistical assumption based on the worker policy and the world model. This avoids additional variance in the manager training what would be caused by the variable performance of individual worker trajectories. As the worker policy becomes more directed and less random over time, the statistical assumption used in the manager training becomes more accurate the better the worker becomes at goal navigation. 
Replacing the actual worker results with a statistical assumption mitigates the impact of the distribution shift that occurs over time due to the worker steadily improving at reaching proposed goals, consequently changing the value of the manager actions over the course of training. Finally, a dilated LSTM variant that is, compared to a regular LSTM, more capable at capturing long-term dependencies further improves the performance of the proposed algorithm. In contrast to our approach, \citet{vezhnevets2017feudal} do not learn any world model and use a statistical assumption for the lower level GSA while training the higher level RMA. As our temporal abstraction process is independent of any agent performance, we by design avoid any distribution shift that may inhibit learning of the higher levels.

\citet{nachum2018data} present HIRO, a model-free HRL algorithm. In HIRO, a manager proposes goals as navigation targets to the worker, whereas the goals are equivalent to the actions of the manager. The reward for the worker is related to the similarity of the current and the goal observations and the goals presented to the worker are relative. The authors find that using raw observations for the similarity calculation works better compared to observation embeddings. In this setting, a moving target problem emerges where the low level policies improve their goal navigation capabilities over time and by that change the effect and reward of high level actions. As a consequence, the amount of required training data increases, which HIRO combats via the use of training data relabeling akin to hindsight experience replay \citep{andrychowicz2018hindsight}. In contract to HIRO, our work separates high level actions from low level goals to avoid the complex exploration of extremely high dimensional action spaces. Additionally, our temporal abstraction process does not rely on lower level agent behaviour to generate high level rewards, thereby side stepping a distribution shift scenario and removing the need for related correction steps. Furthermore, we use hierarchical learned world models to improve data efficiency. In contrast to \citet{nachum2018data}, we do not find the observation to always be the best choice for computing the goal reward.


\citet{pertsch2020long} devise a hierarchical trajectory prediction algorithm called ``Goal-Conditioned Predictor'' (GCP) that can produce observations lying in between a start and a goal observation. By generating one observation exactly in the middle of a start and a goal observation, GCP can be used repetitively in a tree-like hierarchical fashion to iteratively fill in intermediate observations until a complete trajectory is generated. To use GCP for control, a cost function that estimates cost between a start and an end observation is trained using environment interaction data. Finally, an inverse dynamics model is trained to obtain actions from the sequence of generated observations. \citet{pertsch2020long} combine GCP with the Cross-Entropy Method (CEM) \citep{rubinstein1997optimization} to perform model predictive control and note that the hierarchical optimisation scheme results in short sequence chunks per optimization procedure, which works well together with CEM. While GCP has the advantage of reaching the given goal with high probability, it needs a goal to begin with and sufficiently diverse training data. By that, \citet{pertsch2020long} exclude the exploration-exploitation dilemma from their investigation.

\cite{hafner2022deep} extend their MBRL Dreamer \citep{hafner2022mastering} algorithm to a hierarchical setting, called Director, where a manager proposes latent goals to a worker. The concept is similar to the method proposed by \citet{vezhnevets2017feudal}, although here the manager obtains the true reward of the worker instead of a statistical surrogate and both are trained in a learned world model. The authors note that since the manager's action space, i.e.\ the worker's goal space, consists of all possible latent states of the world model, finding good and meaningful goals can be difficult. To alleviate this problem, they modify the manager's action space to be a mixture of categoricals and introduce a goal autoencoder with a mixture of categorical latents. The autoencoder is trained on latent world model states and used to translate manager actions to goals for the worker. The authors find that using this trick improves generalisation capabilities of the manager and effectively reduces the search space for finding achievable and rewarding goals for the worker. Furthermore, the reconstruction error of the goal autoencoder is used as an additional exploration reward bonus, since a high reconstruction error can be an indicator of novel and unseen model states.
While being an HMBRL approach, Director learns only a fine grained one-step world model and does not explore the potential of hierarchical world models. In contrast to our work and similar to other goal-based approaches, Director uses high level manager actions as goals for the low level workers. Contrary to that, our approach learns low dimensional abstract actions independently of the goal dimensions and by design always learns valid goals. Thus, it does not require the countermeasures implemented in \citep{hafner2022deep}.

\citet{mcinroe2023learning} learn a stack of world models that operate at increasing temporal resolution and train a modified SAC \citep{haarnoja2018soft} agent in them.
They name their method ``hierarchical k-step latent'' (HKSL) and explicitly deem the learing of a hierarchy of world models to be an auxiliary task to improve performance on regular RL environments.
Unlike most other H(MB)RL approaches, decision making is still done at the highest temporal resolution of the original environment.
For this, each level's model is equipped with a separate critic and a single actor combines information from all levels to make more foresighted decisions compared to the non-hierarchical setting.
The focus of HKSL lies on improved sample efficiency and performance. In contrast to the majority of other HRL algorithms, HKSL is not concerned with abstract state/action discovery or skill transfer.

\subsection{Semantic H(MB)RL}


\citet{florensa2017stochastic} propose to learn a set of skills in a pre-training phase while using the same neural network policy for the individual skill policies to facilitate synergies during the learning process. The skills are subsequently used by a hierarchical policy in downstream tasks that feature the same dynamics as the pre-training tasks but different reward distributions. Specifically, the reward distributions of the pre-training tasks are chosen with minimal domain knowledge to avoid spoiling the actual task during pre-training. In contrast to our work, no world model is learned and skill activation is controlled by the hierarchical agent at a per time step resolution. Thus, while sub-goal navigation is not necessary, \citet{florensa2017stochastic} need additional machinery to decide when to start and terminate individual skill policies.


\citet{xie2021latent} combine hierarchical online planning using the crossentropy method (CEM) with policy-based low level skill learning in a method they call ``Learning Skills for Planning'' (LSP). To increase efficiency, a world model is used for training with simulated environment interactions. The authors find that a random shooting algorithm like CEM is beneficial for generalisation, as CEM re-starts with a random uniform action distribution at every planning step. In contrast to that, a learned policy can get stuck in a previously learned action distribution when confronted with a change in the sub-task distribution. However, as CEM is computationally expensive in large action spaces, the authors opt for learning skill-specific policies in the high dimensional action spaces and selectively activate them via CEM online-planning. By that, they combine the benefits of both worlds. 
In contrast to our work, LSP does not learn a hierarchy of world models. Furthermore, as the abstract actions in LSP are equivalent to low level skills, CEM becomes increasingly costly the more low level skills are learned. Thus, LSP's applicability to more complex scenarios that require larger numbers of learned skills remains limited.

\citet{singh1992reinforcement} presents a hierarchical DYNA \citep{sutton1991dyna} variant that learns hierarchical world models in MDPs with a factorised structure. The action space of higher level models is the activation of policies trained on lower level models, which can be interpreted as individual skills. Although the hierarchical algorithm performs favourably compared to a regular DYNA agent, in contrast to our work the world model abstraction is handcrafted, requires pretraining of individual skills, and relies on discrete states and actions.

\citet{li2017efficient} present context sensitive RL (CSRL), an HMBRL approach designed for discrete state and action spaces. CSRL learns separate and unique dynamics models, called fragments, for individual features of the environment states. The fragments can be combined to simulate various tasks and by improving performance on one fragment, all tasks that share it profit simultaneously. Thus, CSRL performs temporal abstraction on task level instead of relying on a fixed step size. However, fragment detection needs discrete state and action spaces and requires exhaustive and combinatorially expensive sweeps through both of them.

\citet{gumbsch2024learning} present Temporal Hierarchies from Invariant Context Kernels (THICK), a hierarchical adaption of \citep{hafner2022deep}. The heart of the approach is the GateL0RD RNN cell \citep{gumbsch2022sparsely} which takes the place of the RSSM's regular GRU cell. GateL0RD RNNs use a sparsely changing context variable specialised to represent latent causes that influence a time dependant process in the background. THICK uses a hierarchical RSSM where a higher level world model is trained to predict the slowly varying context variable of the GateL0RD cell. The higher level model is controlled through discrete actions that are learned by an abstract policy. The abstract policy is trained in as self-organised manner to find abstract actions that best express the observed context changes in the GateL0RD cell of the lower level. In contrast to our work, THICK incorporates the hierarchical world model information during agent training via a separate critic on level 0 that evaluates the abstract world model predictions and thus provides more far sighted information to an otherwise flat level 0 policy. THICK is tested in a hierarchical planning scenario using model predictive control as well, which is conceptually similar to our setup. However, in that setting no actor-critic agents are trained and policies are entirely created on the fly via trajectory optimisation techniques.

\section{Conclusion}

In this work we present a goal-conditioned approach for crafting a hierarchy of world models for HRL (herarchical reinforcement learning). Therein, a separate world model per level is constructed, with each level getting progressively coarser in time. Agents are trained via simulations inside the world models, whereas each level houses an agent trained on maximising episode return and another one that is trained to navigate to given goals. To connect the hierarchies, on all levels above the lowest one the world models output goals which are equivalent to lower level world model states. Specifically, the goals on level $l$ can be followed by the specialised agents on level $l-1$ to implement hierarchical decision making. Importantly, the action dimension of each world model level is independent of the goal dimension and abstract actions are produced by processing ground truth environment actions. This way, we obtain relatively low dimensional abstract actions and avoid the problem of having to explore high dimensional action spaces on abstract levels of the model. Additionally, our abstraction process always yields valid goals for lower levels and the abstracted reward is independent of  agent performance and instead based on ground truth data, which circumvents a moving target problem that would otherwise complicate learning on higher levels \citep{vezhnevets2017feudal, nachum2018data}.

In our experiments, the hierarchical world models reliably captured task dynamics with increasingly coarser temporal resolution. However, in its current form our HMBRL approach is at most on par with its non-hierarchical counterpart in terms of learning speed and final performance. We found model exploitation on the abstract level of the hierarchical world model to be a likely explanation for this. More importantly, our findings imply model exploitation to be a potential problem in other HMBRL approaches that feature a similar structure to ours, particularly when abstract actions are produced from low level action sequences. Nonetheless, depending on the environment our approach performed on par with the non-hierarchical baseline while compressing e.g.\ an 8 step sequence of 2D actions to a single 2D or 4D abstract action. This suggests that our abstraction process in conjunction with the hierarchical world model is capable of producing useful abstractions of the environment dynamics without manual intervention. This is in stark contrast to related H(MB)RL approaches that need handcrafted abstractions, use abstract actions with hundreds of dimensions, or are limited to specific problem domains like tabular environment dynamics.

Our algorithm implements goal-conditioned HMBRL with a fixed amount of steps between each goal. While this is a common approach in H(MB)RL, it achieves the hierarchical decomposition of decision making predominantly through temporal abstraction that may completely ignore the semantics of the environment. Being an arguably harder problem to solve, variable length temporal abstraction opens the opportunity to better reflect the semantics of the environment in the process. One example is an abstract action that moves an agent which is located inside a room to the door of the room, independently of how far away it currently is from the agent. This does not only reflect a meaningful semantic abstraction of the environment in terms of rooms an their entrances or exits, but it also opens up the path for specialised low level policies that can be re-used in environments exhibiting similar structure, i.e.\ doors and rooms. Incorporating variable length temporal abstraction into our approach would primarily require a mechanism to detect favourable chunk borders during the temporal abstraction and an alternative action abstraction method that is compatible with variable length sequences. As previously mentioned, the chunking mechanism should be motivated by meaningful or semantic changes in the environment, which is an ambitious requirement and one of the main problems explored by contemporary H(MB)RL research. An action abstraction process capable of handling variable length sequences could be achieved through padding of action sequences to make them compatible with the methods we presented in \ref{sec:temporal_abstraction}. Alternatively, a dedicated RNN could be used to generate an abstract action of a fixed size from an arbitrary sequence of actions.

For future research directions, we consider the model exploitation issue the most pressing aspect. One option, inspired by our investigation of discrete action spaces for the abstract world model levels, is to explore alternative abstraction methods that reduce or prevent model exploitability. Interestingly, the work of \citet{gumbsch2024learning} demonstrates the feasibility of discrete abstract actions if the abstraction process features variable length trajectory chunks. We hypothesise that allowing variable length chunks may provide the opportunity to identify short reoccurring action sequences or those that change the environment context in the same way and label them with a particular abstract action. This avoids wasting the limited budget of discrete abstract actions on a potentially large number of very similar primitive action sequences only since those sequences necessarily have to have be of the same length. In the approach \citet{gumbsch2024learning} present, the context of the GateL0RD RNN cell \citep{gumbsch2022sparsely} is used to detect semantic changes in the environment and therefore useful chunk borders for learning abstract actions, which could result in clearer and higher quality trajectory abstraction. In contrast to that, our current approach performs the abstraction process based on equally sized chunks independently of trajectory semantics. This may require the abstract world model to differentiate between a large number of similar but not equal action sequences that are mainly dictated by the chunking process. In that setting, distinct abstract actions could potentially lack expressive diversity in contrast to continuous ones, which explains our experimental findings (c.f. Section~\ref{sec:experimens_and_discussion}).

Another option is to investigate the applicability of techniques specifically designed to alleviate model exploitation, perhaps from the offline RL community, where the problem has been a central concern for a long time.

\bibliography{main}

\end{document}